# A Hybrid LP–RPG Heuristic for Modelling Numeric Resource Flows in Planning

**Amanda Coles**                                    AMANDA.COLES@KCL.AC.UK
**Andrew Coles**                                    ANDREW.COLES@KCL.AC.UK
**Maria Fox**                                       MARIA.FOX@KCL.AC.UK
**Derek Long**                                      DEREK.LONG@KCL.AC.UK
*Department of Informatics*
*King's College London*
*Strand Building*
*London, WC2R 2LS, UK*

## Abstract

Although the use of metric fluents is fundamental to many practical planning problems, the study of heuristics to support fully automated planners working with these fluents remains relatively unexplored. The most widely used heuristic is the relaxation of metric fluents into interval-valued variables — an idea first proposed a decade ago. Other heuristics depend on domain encodings that supply additional information about fluents, such as capacity constraints or other resource-related annotations.

A particular challenge to these approaches is in handling interactions between metric fluents that represent exchange, such as the transformation of quantities of raw materials into quantities of processed goods, or trading of money for materials. The usual relaxation of metric fluents is often very poor in these situations, since it does not recognise that resources, once spent, are no longer available to be spent again.

We present a heuristic for numeric planning problems building on the propositional relaxed planning graph, but using a mathematical program for numeric reasoning. We define a class of producer–consumer planning problems and demonstrate how the numeric constraints in these can be modelled in a mixed integer program (MIP). This MIP is then combined with a metric Relaxed Planning Graph (RPG) heuristic to produce an integrated hybrid heuristic. The MIP tracks resource use more accurately than the usual relaxation, but relaxes the ordering of actions, while the RPG captures the causal propositional aspects of the problem. We discuss how these two components interact to produce a single unified heuristic and go on to explore how further numeric features of planning problems can be integrated into the MIP. We show that encoding a limited subset of the propositional problem to augment the MIP can yield more accurate guidance, partly by exploiting structure such as propositional landmarks and propositional resources. Our results show that the use of this heuristic enhances scalability on problems where numeric resource interaction is key in finding a solution.

## 1. Introduction

Domain-independent planning research in the last decade has focussed, for the most part, on propositional planning, leading to important discoveries and powerful new heuristics for planning in propositional domains. Relatively little effort has been invested in planning with metric fluents, despite their importance in representing many practical planning problems.

Numbers are essential for efficiently encoding resources, such as money, fuel and materials. The state-of-the-art remains the influential approach proposed by Hoffmann (2003), which extends the ignore-delete-effects relaxation from propositional fluents to metric fluents by tracking the accumulating upper bound on increasing fluents and ignoring decreasing (negative) effects. A symmetrical treatment of the fluents for the purposes of determining a lower bound leads to a representation which is equivalent to an interval for each metric fluent, with goals and preconditions being satisfied provided that *some* value in the interval is sufficient to satisfy each condition. Although LPG (Gerevini, Saetti, & Serina, 2006) and MIPS (Edelkamp, 2003) are both capable of handling metric fluents, both depend on the same relaxation heuristic to offer search guidance. Other approaches have been explored (Do & Kambhampati, 2001; Koehler, 1998), but have been comparatively less successful.

Planners using the MetricFF heuristic are generally not very effective at solving problems in which there are complex interactions between the values of numeric resources, such as the exchange of quantities of one or more materials for the production of others. In contrast, solving problems with numbers is at the heart of operational research and mathematical programming techniques. Many powerful solvers have been developed for solving Linear Programming problems (LPs) and Mixed Integer Programming problems (MIPs), in which problems are expressed as linear constraints over variables (which, in the case of MIPs, can be integers). Although there have been efforts to exploit linear programming in propositional planning, either to schedule actions (Long & Fox, 2003a) or directly, as a heuristic (van den Briel, Benton, Kambhampati, & Vossen, 2007), relatively little work has considered the exploitation of linear programming techniques to improve the behaviour of *numeric* domain-independent planners (Kautz & Walser, 2000; Shin & Davis, 2005; Wolfman & Weld, 2000; Benton, van den Briel, & Kambhampati, 2007) (this work is further considered in Section 1.1).

In this paper, we revisit the issue of planning with numeric resources, beginning with the Metric Relaxed Planning Graph (RPG) heuristic (Hoffmann, 2003). We focus specifically on domains that exclusively exhibit what we call producer-consumer behaviour (defined in Section 2.3), in which actions increase or decrease numeric resources by fixed quantities. Of course, this represents only a subset of possible numeric behaviours, but it is a common and intuitive one. Furthermore, it is easy to recognise syntactically in a domain encoding, so it is simple to resort to alternative strategies for domains that do not conform to this constraint, which might include the use of producer-consumer relaxations or approximations of domains with more complex numeric behaviour.

We explore the behaviour of the RPG heuristic, demonstrating how the very typical patterns of interactions in producer-consumer numeric planning domains can lead to highly uninformative heuristic guidance, particularly when domains offer opportunities for exchanges between metric variables. To address this, we introduce a novel heuristic based on a mixed integer program (MIP), used alongside the RPG, to better capture numeric constraints. Having described how the MIP is constructed, and how it can be used to complement the RPG, we discuss extensions of it to improve identified weakness, and also to encode more information about the propositional behaviour of the problem. We evaluate the LP-RPG heuristic by exploring the spectrum between, at one end, a strict separation of numbers and propositions into the MIP and RPG components; and, on the other, discarding the RPG entirely and encoding the preconditions and effects of actions entirely as a MIP. In doing so, we will determine where the best trade-off between the two lies.

The work we report in this paper is an extension of our earlier work reporting development of LP-RPG (Coles, Fox, Long, & Smith, 2008). It extends that work both with additional detail and with several variants of the core heuristic, exploring the impact of tighter integration between the propositional and metric fluents in the heuristic.

## 1.1 Related Work

The integration of linear programming (LP) or MIP techniques into planning has been considered in a number of contexts. The most relevant to the present work is the use of an LP as the basis of a heuristic for propositional over-subscription planning problems (Benton, Do, & Kambhampati, 2005). In this setting, the goal of planning is to find a plan with maximum utility, defined in terms of the reward for the goals achieved, minus the costs occurred in achieving them. Benton et al. use the LP as an optimisation tool to help to decide which set of goals the planner should satisfy in order to achieve maximum reward. Both the work of Benton et al. and the work described in this paper exploit a relaxation of the action ordering rather than of their effects, and both employ an LP as well as an RPG structure. The two key differences are that the focus in this work is on using the MIP to capture interactions within *numeric* planning problems, and we will rely on a conventional Relaxed Planning Graph (Hoffmann & Nebel, 2001) for propositional reasoning, rather than also encoding this structure in the MIP. The work of van den Briel et al. (van den Briel et al., 2007; van den Briel, Vossen, & Kambhampati, 2008) also explores the use of mathematical programming to encode and solve planning problems.

The structure in the LP and MIP models proposed by Benton et al. (2007) and van den Briel et al. (2008) makes them time consuming to solve, requiring actions to be selected to satisfy preconditions and effects of other actions, and delete effects to be paired with add effects. In contrast, the MIP and LP models we propose do not attempt to capture most of the causal plan structure, making the construction of solutions to our programs at each state much more feasible. A further difference is in the interaction between the two components: the integration between the MIP and the RPG in LP-RPG is much tighter, with the MIP being used in graph building to indicate variable bounds, and in relaxed plan extraction to indicate the actions to use. By comparison, in the earlier approach, the MIP is used solely to introduce a bias into RPG action selection, giving preference to actions used in the solution of the MIP.

Linear programming has been exploited in planning in other work. LPSAT (Wolfman & Weld, 2000) uses a planning-as-satisfiability approach, linked to the use on an LP solver to ensure that literals representing (linear) constraints on metric fluents are maintained during the plan construction. There is no heuristic guidance in the search, which is based on a standard DPLL search for a satisfying assignment combined with confirmation that the corresponding LP is satisfiable. IP-SAT (Kautz & Walser, 2000) uses a MIP encoding of planning problems as the basis for solving them, as do Vossen et al. (1999), in a similar way to the later work of van den Briel et al. (van den Briel et al., 2008). In these planners the MIP is used directly at the heart of the solver, with planning problems being translated into MIPs rather than being used to guide the search. TM-LPSAT (Shin & Davis, 2005) uses the LPSAT system to solve planning problems with continuous processes. Kongming (Li & Williams, 2008) is another example of a planner that exploits compilation of planning problems into mathematical programs, solved using CPLEX, to tackle hybrid mixed-continuous planning problems. Our own COLIN system (Coles, Coles, Fox, & Long, 2012) also uses LP encodings to manage reasoning about the effects of continuous processes.

A completely different use of linear and mixed integer programming in planning lies in work by Ono and Williams (2008) and also Blackmore, Ono and Williams (2011) which uses mixed integer programming as the foundation for solving the problem of risk allocation in plan-level control systems.

Alternative approaches to handling numeric variables in planning include those implemented in MetricFF, discussed in detail in Section 3, Sapa (Do & Kambhampati, 2001) and Resource-IPP (Koehler, 1998). In Sapa heuristic cost estimates generated using relaxed plan extraction are supplemented with additional costs representing the minimal set of additional resource producing actions required to achieve the resource requirements of the relaxed plans. This approach is straightforward to implement and is an interesting modification of the pure relaxed plan heuristic, but it separates the problem of producing resources from the solution of the rest of the problem, with the consequence that a relaxed plan using a few steps with high resource demands will be constructed in preference to a longer plan with lower demands. The heuristic value of the state will then be distorted by the penalty attached to the relaxed plan to achieve its high resource requirement, potentially hugely overestimating the true distance of the state to the goal. Resource-IPP depends on the identification of consumers and producers, as we do in this paper, and it then builds a resource time map that tracks the production and consumption of the resources during a Graphplan-based search for a plan. The approach leads to an extension of the mutex relation that is used to constrain the search in Graphplan (Blum & Furst, 1995). However, the iterative-deepening search used in Graphplan-based planners is not scalable to solve large problems and forward state-space search has proved a dominant strategy in the past decade.

## 2. Problem Definition

In this section we define the class of planning problems that we will consider in this paper. They are a subset of the general class of PDDL 2.1 non-temporal, numeric planning problems, which represent linear producer–consumer problems. We include as an example the Settlers domain, which will then be used throughout the paper to illustrate the ideas presented.

### 2.1 PDDL 2.1 Numeric Planning Problems

In this work, we are concerned with finding sequential plans to solve non-temporal, numeric planning problems, as defined using (a subset of) PDDL 2.1 (Fox & Long, 2003). Within PDDL, this class of problems can be defined as follows:[1]

---

1. PDDL2.1 also supports the specification of an objective function to measure plan quality, defined over the numeric variables in the planning problem, but in this work we focus only on minimising plan length.

- $I$ is the initial state, where a *state* consists of a set of propositions, and/or an assignment of values to a set of numeric variables. For notational convenience, we refer to the vector of numeric values in a given state as $\mathbf{v}$ and the propositional facts as $F$.

- $A$, a set of actions. Each $a \in A$ is a tuple $\langle pre, eff \rangle$:

  - $pre$ the preconditions of $a$: these conditions must hold in the state in which $a$ is to be executed.

  - $eff$ the effects of $a$: when $a$ is applied, the state is updated according to these effects. $eff$ consists of:

    * $eff^-$, propositions to be deleted from the state;
    * $eff^+$, propositions to added to the state;
    * $eff^n$, effects acting upon numeric variables.

- $G$, a goal: a set of propositions $F^\star$ and a set of conditions over numeric variables, $N^\star$. Each of these sets may be empty. A state $\langle F, \mathbf{v} \rangle$ is a goal state if $F^\star \subseteq F$ and $\mathbf{v}$ satisfies each condition in $N^\star$.

In the general case, PDDL numeric conditions (as used in $pre$ and $N^\star$) are expressed in the form:

$$\langle f(\mathbf{v}), op, c \rangle \qquad \text{s.t.} \qquad op \in \{\leq, <, =, >, \geq\},\ c \in \Re$$

Numeric effects (as in $eff$n) are expressed as:

$$\langle v, op, f(\mathbf{v}) \rangle \qquad \text{s.t.} \qquad op \in \{\times=, +=, =, -=, \div=\}$$

In common with Hoffmann's work on MetricFF (2003) we restrict our attention to preconditions that can be expressed in Linear Normal Form (LNF). That is, the expression $f(\mathbf{v})$ within preconditions must be in the form of a weighted sum of the state variables plus a constant, $\mathbf{w}.\mathbf{v} + k$. Likewise, we consider only numeric effects where $f(\mathbf{v})$ is in LNF, and $op \in \{+=, =, -=\}$. These restrictions guarantee termination in the construction of the RPG when evaluating a state: introducing non-LNF preconditions, or scaling effects, can lead to asymptotic numeric behaviour where certain conditions are only satisfied at an infinite limit. For the LP-RPG heuristic we describe in this work, we further require that the numeric behaviour of actions can be represented as producer–consumer behaviour. That is, all effects cause constant increments or decrements to the variables they affect and, apart from in specific circumstances, we do not permit assignment effects. We will precisely define these notions and the circumstances in which we allow assignment effects later in the paper.

A solution to a planning problem is a (sequential) *plan*: a sequence of actions that transforms the initial state into a goal state, respecting all preconditions on action application. In a state $\langle F, \mathbf{v} \rangle$, the application of an action with effects $eff^-$, $eff^+$, $eff^n$ yields a successor state $\langle F', \mathbf{v}' \rangle$, where:

$$F' = (F \setminus eff^-) \cup eff^+$$

$$
\begin{aligned}
\mathbf{v}'[x] \quad & op\ (\mathbf{w}.\mathbf{v} + c) && \text{if } \exists \langle v, op, \mathbf{w}.\mathbf{v} + c \rangle \in eff^n \\
\mathbf{v}'[x] \quad & = \mathbf{v}[x] && \text{otherwise}
\end{aligned}
$$

## 2.2 An Example Problem: Settlers

The Settlers domain, introduced in the 2002 International Planning Competition (IPC) (Long & Fox, 2003b) and used again in 2004 (Hoffmann & Edelkamp, 2005), is a good example of a problem exhibiting interesting use of metric fluents. The aim in Settlers problems is to build up transport and building infrastructure through the extraction, refinement and transportation of materials. The numeric structure of the domain is perhaps the most sophisticated of the IPC domains to date. First, there are six numeric resources and several actions that act upon each. The available resources, and the effects of actions upon them (consumption of a resource is shown as a negative value and production is shown as a positive value) are shown in Table 1[2]. Another interesting feature of this domain is that not all resources can be directly produced: whilst the raw materials Timber, Stone and Ore can be directly extracted, Wood, Coal and Iron must be *refined* from their respective raw form. Finally, the domain contains *transferable* resources. In addition to the actions shown in the table, by which resources can be refined or consumed to fuel transportation, resources can be loaded and unloaded from vehicles. The effect of such load and unload actions is to increase or decrease the amount of a resource on a vehicle, and decrease or increase the amount stored at a given location. Apart from consuming or 'producing' (i.e. releasing) the remaining cargo space of the vehicle, no resource is produced or consumed during loading and unloading — it is only moved. However, expressing the model in PDDL requires the pair of effects described, decreasing one variable and increasing another, which is indistinguishable from a combination of production and consumption.

## 2.3 Producer–Consumer Problems

We now define the constrained *producer–consumer* numeric behaviour considered in this paper. We first define *producer* or *consumer* actions, with two categories of producer. Using these we then define the notion of a *producer–consumer variable*. The identification of consumers and producers is not a new idea — it is common to identify resource producers and consumers in scheduling (for example, Laborie does so in his work on scheduling with resource constraints, see Laborie, 2003).

### 2.3.1 Producer–Consumer Actions

A simple production action is defined as follows:

---

**Definition 2.1  — Simple Producer**

A ground action $a$ is a simple producer of a given numeric variable $v$ *iff*:

- it has an effect (`increase (v) c`) (where $c$ is a positive constant) and

- it has no precondition that refers to $v$.

---

This definition has two important consequences:

---

| Action | Timber | Stone | Ore | Wood | Coal | Iron |
|---|---|---|---|---|---|---|
| Move cart | | | | | | |
| Move train | | | | | -1 | |
| Move ship | | | | | -2 | |
| Fell timber | +1 | | | | | |
| Quarry stone | | +1 | | | | |
| Mine ore | | | +1 | | | |
| Saw wood | -1 | | | +1 | | |
| Make coal | -1 | | | | +1 | |
| Smelt iron | | | -1 | | -2 | +1 |
| Build cabin | | | | | | |
| Build quarry | | | | | | |
| Build mine | | | | -2 | | |
| Build saw-mill | -2 | | | | | |
| Build iron-works | | -2 | | -2 | | |
| Build coal-stack | -1 | | | | | |
| Build dock | | -2 | | -2 | | |
| Build wharf | | -2 | | | | -2 |
| Build house | | -1 | | -1 | | |
| Build cart | -1 | | | | | |
| Build train | | | | | | -2 |
| Build ship | | | | | | -4 |
| Build rail | | | | -1 | | -1 |

Table 1: Production and consumption in the Settlers domain

1. A simple producer produces uniformly: if a state $s$ satisfies its preconditions, then the effect by $a$ upon $v$ is always to increase its value by the same constant amount, $c$, irrespective of the precise details of $s$.

2. The potential maximum value of $v$ that can be attained through the use of $a$ is not restricted by the value of $v$ itself: there are no minimum or maximum bounds on $v$ that must hold to allow production.

We define a *bounded* producer as follows:

---

**Definition 2.2  — Bounded Producer**

A ground action $a$ is a bounded producer of a numeric variable $v$ *iff*:

- it has an effect (`increase (v) c`) (where $c$ is a positive constant),

- it has a precondition (`<= (v) (- ub c)`) and

- it has no other preconditions depending on `v`.

---

A bounded producer, $a$, can only be applied if $v \leq (ub - c)$. Therefore, the maximum amount of $v$ that can be attained using $a$, denoted $max\_prod(a, v)$, is $ub$, achieved by applying $a$ in a state where $v = (ub - c)$. (In practice, such a state might not be reachable and the actual upper bound on the value of $v$ reachable using $a$ might be lower than $ub$). For an simple producer, $a'$, we assume $max\_prod(a', v) = \infty$.

We define a bounded consumer as follows:

---

**Definition 2.3  — Bounded Consumer**

A ground action $a$ is a consumer with respect to a given numeric variable $v$ *iff*:

- it has a precondition (`>= (v) (+ lb c)))`,

- it has an effect (`decrease (v) c`), where $c$ is a constant and

- it has no other preconditions depending on `v`.

---

This definition is analogous to the bounded producer, since it requires that $v$ exceed a minimum value before allowing consumption. As a consequence, $lb$ is the minimum amount of $v$ that can be attained using $a$, denoted $min\_cons(a, v)$ (by applying $a$ in a state where $v = lb + c$).

There are, of course, many other resource use behaviours that might be encoded in planning domains. The producer-consumer behaviour we identify is a natural and intuitive one (a producer produces a fixed quantity of a resource and a consumer consumes a fixed quantity and depends on the availability of that quantity). There are variants that can be compiled into this form (e.g. consumers that must leave a fixed sized store of resource untouched simply translate the origin of the resource measurement) and we also consider, below, other possible extensions of this basic behaviour. Nonetheless, we must emphasise that the heuristic we develop in this paper is targeted at this producer-consumer behaviour and its usefulness depends on how common such domains are in practice. The frequent occurrence of this model in scheduling with resources suggests that it is a natural and useful behaviour.

### 2.3.2 PRODUCER–CONSUMER VARIABLES

With these definitions of (bounded) producer and consumer actions, we define properties of the variables that they manipulate:

---

**Definition 2.4  — Producer–Consumer Variable**

A variable $v$ denotes a resource that is produced/consumed iff:

- the set $prod(v)$ of actions that increase the value of $v$ contains only bounded producers,

- the set $cons(v)$ of actions that decrease the value of $v$ contains only consumers,

- the upper bound on $v$ is the same in all bounded producers for $v$ and

- the lower bound on $v$ is the same in all consumers for $v$.

---

### 2.3.3 Handling Integer Resources

Consumer actions (Definition 2.3) each require that the amount of a resource available for consumption must be at least as much as the consumer actually consumes. In some domain encodings, behaviour that is essentially consistent with producer–consumer patterns is represented using a precondition on a consumer action that consumes $c$ units of $v$ in the form $v > k$ rather than $v \geq c$ (and $k < c$). In the general case, where $v \in \Re$, all we know is that if $v \geq k + \epsilon$ (where epsilon is infinitesimal and positive), $c$ units of $v$ can be consumed, suggesting a lower bound on $v$ of $(k - c + \epsilon)$. However, in the case that all consumers consume integral quantities, we can rewrite the strict inequality since $\epsilon$ must be 1 in this case. Consider, for example, this fragment of a load action from Settlers:

```
:precondition (> (available timber l1) 0)
:effect       (decrease (available timber l1) 1)
```

Because the effects that change the quantities of available resources are integral, this can be rewritten:

```
:precondition (>= (available timber l1) 1)
:effect       (decrease (available timber l1) 1)
```

A similar transformation can be used when all the constant effects on a variable are rational, simply by finding the least common multiple, $LCM$, of the denominators of the fractions involved and using $\epsilon = 1/LCM$.

## 3. MetricFF Revisited

In this section we briefly review the way in which MetricFF (Hoffmann, 2003) handles metric fluents and highlight some of the weaknesses in this approach when faced with particular kinds of numeric behaviours in planning domains.

### 3.1 The Metric Relaxed Planning Graph Heuristic

The Metric RPG heuristic is based on performing a relaxed reachability analysis forwards from the state to be evaluated, where the reachability analysis is captured in a planning graph (Blum & Furst, 1995) structure. Two elements of the domain are relaxed: delete conditions of actions are ignored and optimistic upper and lower bounds are used to record the interval of possible values that a metric fluent may reach. Positive effects on a metric variable increase the upper bound on its reachable values and negative effects decrease its lower bound. Satisfaction of preconditions is tested by checking some value in the interval of each variable satisfies each metric condition in the precondition. It is interesting to note that preconditions are tested individually, so it is possible, in principle, that no single value could satisfy all the conditions simultaneously, even though each condition is separately satisfied by some value. Conjunctions of convex preconditions, which includes linear conditions, will be satisfiable by some value in the case that each condition is satisfiable in the interval associated with a variable, except in the case that the conjunction is inconsistent, which is likely to arise only in erroneous domain encodings.

MetricFF allows preconditions to combine multiple variables, and effects to depend on the values of other variables. In LP-RPG we allow linear combinations of variables in preconditions, but effects must conform to the producer–consumer definitions above and allow only constant increases or decreases.

Heuristic evaluation of a state using the Metric RPG heuristic is undertaken in two phases: the graph expansion phase and the solution extraction phase. We now remind the reader of these two processes for convenience of reference in the discussion that follows.

### 3.1.1 Metric RPG Expansion

Graph expansion can be concisely defined as follows:

---

**Definition 3.1 — RPG Expansion**

Let $F(i)$ denote a *fact layer*, comprising:

- $FP(i)$, a set of propositions;

- $FV(i)$, an array of upper- and lower- bound pairs for each task numeric variable $v$

$A(i)$ denotes an *action layer*, consisting of a list of ground actions. An RPG begins with a fact layer, $F(0)$, defined based on the state $S$ to be evaluated:

- $FP(0)$ contains the propositions that hold in $S$;

- Each entry $\langle LB_v, UB_v \rangle \in FV(0)$ is set to $\langle S[v], S[v] \rangle$, i.e. the value of $v$ in $S$.

The RPG is expanded by adding successive action layers, followed by new fact layers:

- Action layer $A(i+1)$ contains all actions $a \in A$, such that:

  - the propositional preconditions of $a$ are in $FP(i)$;

  - the numeric preconditions of $a$ are satisfied for *some* values of the variables in $FV(i)$.

- Fact layer $FP(i+1)$ is then determined from $A(i+1)$:

  - The propositions $FP(i+1)$ are each of $FP(i)$, plus any new facts added by an action in $A(i+1)$;

  - The values of the numeric values $FV(i+1)$ are first set to $FV(i)$, then updated by extending the interval for each variable to include the values achieved by the maximum and minimum possible assignment effects, for each action $a \in A(i+1)$ in turn.

- Until the termination condition is met, the RPG is expanded with further action-layer–fact-layer pairs.

---

The reachability analysis therefore consists of alternate steps: determining which actions are applicable, and hence instantiating the next action layer, and then using these to extend the next fact layer. This process is presented graphically in Figure 1, for a small problem with facts $f_0...f_n$ and numeric variables $v_0, v_1$. Considering first the propositions:

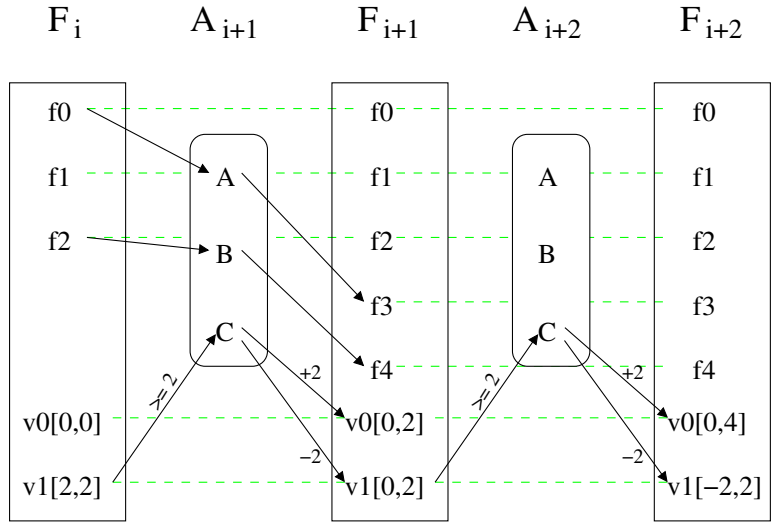

Figure 1: Portion of a relaxed planning graph, where $C$ produces 2 units of $v0$, but consumes 2 units of $v1$

- Arrows from fact to action layers denote the precondition dependencies of actions — for instance, action A can appear in layer $A(i+1)$ because $f_0$ is present in $F(i)$.

- Arrows from action to fact layers denote effects — for instance, $f_3$ is in $F(i+1)$ because it was added by A.

For the numeric variables $v_0, v_1$, the bounds are shown in square brackets. Action $C$ can be seen to have one precondition ($v_1 \geq 2$) and two effects (increase $v_0$ by 2, decrease $v_1$ by 2). $C$ exhibits producer–consumer behaviour — it consumes 2 units of $v_1$, and produces 2 units of $v_0$. Its preconditions are satisfied in $F(i)$ and therefore its effects are applied in layer $A(i+1)$, where the upper bound on $v_0$ has increased and the lower bound on $v_1$ has decreased. Moreover, the bounds change *again* in layer $F(i+2)$, through a further possible application of $C$.

Because variable bounds can continue to diverge in this way, RPG expansion needs a well-defined termination condition. In the positive case, we can terminate with success at the first layer $F(i)$ where all goal propositions are in $FP(i)$ and all goal numeric expressions are satisfied by $FV(i)$ (in the relaxed sense). In the negative case, we terminate with failure at $F(i)$ if all three of the following hold:

1. no actions appear in $A(i+1)$ that were not present in $A(i)$ (and hence no new propositions would be present in $FP(i+1)$),

2. for all hitherto unsatisfied preconditions $v \geq c$, of any action, $UB(v)$ would not change between $FV(i)$ and $FV(i+1)$, and

3. for all hitherto unsatisfied preconditions $v \leq c$, of any action, $LB(v)$ would not change between $FV(i)$ and $FV(i+1)$.

---

**Algorithm 1**: Adding an action to a relaxed plan

> **Data**: $R$ — a metric RPG, $a$ — an action to include in the relaxed plan, $q$ — the subgoal
>     queue

**1** **foreach** *propositional precondition pre of a* **do**
**2**     $l \leftarrow$ layer at which *pre* first appears;
**3**     **if** $l > 0$ **then** insert *pre* into $q[l].prop$;

**4** **foreach** *numeric precondition pre of a* **do**
**5**     $l \leftarrow$ layer at which *pre* first appears;
**6**     **if** $l > 0$ **then** insert *pre* into $q[l].num$;

---

The intuition behind these conditions is that the monotonic expansion of the RPG implies that, if no new facts are appearing and no more numeric preconditions could become satisfied at a future layer, graph expansion has stagnated and the relaxed problem is unsolvable.

### 3.1.2 METRIC RPG SOLUTION EXTRACTION

Having expanded the planning graph and found that a relaxed solution exists (all the goals have appeared), the next step is to extract a relaxed solution plan. This is done by regressing through the planning graph, using a priority queue of intermediate sub-goals (latest layer first). For each subgoal, an achieving action is added to the relaxed plan and its preconditions are added to the queue as goals to be achieved at an earlier layer.

The relaxed plan extraction algorithm is shown in Algorithm 2. In lines 3–6, the priority queue is initialised with the top-level goals of the problem. Both propositional and numeric goals are added to the priority queue to be achieved in the earliest fact layer in which they appeared. Once the priority queue is seeded, solution extraction proceeds by regressing layer-by-layer. For propositions, it suffices to find an action that adds the fact and then to increment the heuristic value by one and add the preconditions of the achieving action to the queue, using Algorithm 1. For numeric preconditions, the process is slightly more involved:

- If the subgoal is to achieve $v \geq k$ or $v \leq k$ at layer $F(l)$, and there is an action in $A(l)$ that assigns the value of $k$ to $v$, then that action is chosen to satisfy the subgoal.

- Otherwise, if $v \geq k$ (or $v \leq k$) must be achieved at layer $F(l)$, then actions increasing (decreasing) $v$ are chosen from those in $A(l)$ until the residual value of $k$ (i.e. the original value of $k$ adjusted to take into account the effects of the selected actions) is small enough (large enough) to be reachable in layer $F(l-1)$. The residual condition $v \geq k$ ($v \leq k$) is added to the queue as a subgoal to be achieved in $F(l-1)$, with the modified value of $k$.

Note that at lines 13, 19, 25, 32 and 39, the actions chosen from action layer 1 are recorded by adding them to the set $ha$. These are used as the basis of the helpful action set: any action with an effect in common with the action set $ha$ is considered helpful. Helpful actions are an important element in the performance of MetricFF: the actions that achieve effects that are exploited in the relaxed solution from a state are promoted in the search from that state.

---

**Algorithm 2**: Relaxed plan extraction

---

**Data**: $R$ - a metric RPG; $F^\star$, $N^\star$ - problem goals
**Result**: $ha$ - helpful actions, $h$ - A heuristic value

1  $ha \leftarrow \emptyset$, $h \leftarrow 0$;
2  $q \leftarrow$ deepest-first priority queue of goal layers;
3  **foreach** $p \in F^\star$ **do**
4  $\quad$ $l \leftarrow$ layer at which $p$ first appears;
5  $\quad$ insert $p$ into $q[l].prop$;

6  **foreach** $f \in N^\star$ **do**
7  $\quad$ $l \leftarrow$ layer at which $f$ first holds;
8  $\quad$ insert $f$ into $q[l].num$;

9  **while** $q$ *not empty* **do**
10 $\quad$ $(l, \langle prop, num \rangle) \leftarrow pop(q)$;
11 $\quad$ **foreach** $p \in prop$ **do**
12 $\quad\quad$ $h \leftarrow h + 1$; $a \leftarrow$ an achiever for $p$;
13 $\quad\quad$ **if** $a$ *in action layer 1* **then** add $a$ to $ha$;
14 $\quad\quad$ $prop \leftarrow prop \setminus$ add effects of $a$;
15 $\quad\quad$ call Algorithm 1 with $R, a, q$;

16 $\quad$ **foreach** $(v \geq c) \in num$ **do**
17 $\quad\quad$ **if** *an action* $a \in A(l)$ *assigned* $v = k, k \geq c$ **then**
18 $\quad\quad\quad$ $h \leftarrow h + 1$;
19 $\quad\quad\quad$ **if** $l = 1$ **then** add $a$ to $ha$;
20 $\quad\quad\quad$ call Algorithm 1 with $R, a, q$;
21 $\quad\quad\quad$ remove all $(v \geq c'), c' \leq k$ and $(v \leq c'), c' \geq k$ from $num$;

22 $\quad$ **foreach** $(v \leq c) \in num$ **do**
23 $\quad\quad$ **if** *an action* $a \in A(l)$ *assigned* $v = k, k \leq c$ **then**
24 $\quad\quad\quad$ $h \leftarrow h + 1$;
25 $\quad\quad\quad$ **if** $l = 1$ **then** add $a$ to $ha$;
26 $\quad\quad\quad$ call Algorithm 1 with $R, a, q$;
27 $\quad\quad\quad$ remove all conditions $(v \leq c), c \geq k$ from $num$;

28 $\quad$ **foreach** $(v \geq c) \in num$ **do**
29 $\quad\quad$ **while** $FV(l-1)[v].upper < c$ **do**
30 $\quad\quad\quad$ $h \leftarrow h + 1$; $a \leftarrow$ next increaser of $v$;
31 $\quad\quad\quad$ decrease $c$ by $\delta(v, a)$;
32 $\quad\quad\quad$ **if** $l$ *is 1* **then** add $a$ to $ha$;
33 $\quad\quad\quad$ call Algorithm 1 with $R, a, q$;
34 $\quad\quad$ **if** $l > 0$ **then** insert $(v \geq c)$ into $q[l-1].num$;

35 $\quad$ **foreach** $(v \leq c) \in num$ **do**
36 $\quad\quad$ **while** $FV(l-1)[v].lower > c$ **do**
37 $\quad\quad\quad$ $h \leftarrow h + 1$; $a \leftarrow$ next decreaser of $v$;
38 $\quad\quad\quad$ increase $c$ by $\delta(v, a)$;
39 $\quad\quad\quad$ **if** $l$ *is 1* **then** add $a$ to $ha$;
40 $\quad\quad\quad$ call Algorithm 1 with $R, a, q$;
41 $\quad\quad$ **if** $l > 0$ **then** insert $(v \leq c)$ into $q[l-1].num$;

---

### 3.1.3 USE OF THE RELAXED PLAN DURING SEARCH

The relaxed plan, computed during heuristic calculation, is used in two ways during search. MetricFF makes use of a two-stage search approach, and in both the number of actions in the relaxed plan is used as a heuristic goal-distance estimate. The first search phase, enforced hill-climbing (EHC), is a greedy hill climbing search approach. It can be thought of as performing breadth first search forward from the initial state $I$, with state progression through action application, until a state with new global-best heuristic value is found. When such a state, $S$, is found, all other states are discarded and EHC search continues in the same manner from $S$. This search strategy is incomplete due to its greedy nature, discarding all states other than $S$ could lead to the loss of a solution. As such, it is followed by a complete WA* search, in order to guarantee completeness of the planner (subject to sufficient time and memory).

Since the EHC phase is already incomplete, but designed to find solutions quickly, MetricFF makes use of another completeness-sacrificing technique in order to attempt to guide the planner to solutions more quickly. This technique is referred to as *helpful action pruning*. Here, only actions that are in the helpful action set for each state are considered for successor generation: the actions that are not helpful are discarded. Note this pruning is not used in best-first search as it would compromise completeness. In practice helpful action pruning improves the performance of MetricFF on many domains, however, it can lead to difficulties if the actions needed to find a solution from a given state do not appear in the helpful action set. To attempt to compensate for this, if EHC terminates after considering only helpful actions, it returns to the state with the last global-best heuristic value, and searches again considering *all* applicable actions, until either it terminates once again (leading to WA*) or a state with a new global-best heuristic value is found, at which point helpful-action pruning is re-enabled, and EHC continues.

## 3.2 Problems with the Metric RPG Heuristic

Although the Metric RPG is a powerful tool to support planning with metric fluents, there are some common situations in planning problems in which the heuristic gives very flawed guidance. These problems include *resource persistence* and *cyclical resource transfer*. We discuss how each of these phenomena can result in misleading heuristic guidance through the relaxed plan length poorly approximating the actual solution length and through *helpful action distortion*.

### 3.2.1 RESOURCE PERSISTENCE

Resource persistence is a consequence of using a relaxation that ignores negative effects. When a resource is consumed it does not disappear in the relaxation of negative effects. The opportunity to reuse the resource can suggest that there is a significantly shorter plan available than is the case in reality. Although this problem occurs for both propositional and metric fluents, the fact that metric fluents commonly encode resources that must be carefully managed means that the problem is often more acute in domains with metric fluents. For example, in a state in the Settlers domain in which 2 units of each resource have been produced and no ship is required (either as a goal or as a means of travel to an otherwise inaccessible location), no relaxed plan will require the production of any further resources (see Table 1).

One approach for approximating the number of missing resource production actions was introduced in the planner Sapa (Do & Kambhampati, 2001). If $v = s$ in the state being evaluated, and the relaxed plan consumes $c$ units of $v$, but only produces $p$, then if $(c - p) > s$, there is a shortfall on production of $v$, which would necessitate additional actions being added to the relaxed plan. In this case, if the maximum amount of $v$ that can be produced by a single action is $\Delta v$, the heuristic value is increased by:

$$\left\lceil \frac{c - p - s}{\Delta v} \right\rceil.$$

This increase is a lower bound on the number of additional actions needed and, while it does not indicate what the actions might be, serves to increase the heuristic value of states whose relaxed plans have resource production shortfalls. It does, however, have two main limitations. First, relaxed plan extraction (such as the approach shown in Algorithm 2) chooses actions without consideration for their undesirable resource consumption side-effects. A good search choice might lead to a state with a worse heuristic value purely because of an accident of the choice of achieving actions (consuming resource unnecessarily). Second, by increasing the heuristic value without adding specific additional actions to the relaxed plan, the helpful actions do not record the fact that appropriate resource production is helpful.

### 3.2.2 Cyclical Resource Transfer

The phenomenon of Cyclical Resource Transfer (CRT) is a consequence of the encoding of actions that move resources around, combined with the relaxation of negative effects. To encode movement of a resource, it is removed from one location and added to another. The removal is encoded as a decrease and this is relaxed when building the Metric RPG. As a result, moving resources appears to generate new resource at the destination, making movement a spuriously attractive alternative to production. Consider a state in which 1 unit of timber and a cart are at a location, $p1$, and the goal is to have 2 units of timber at $p1$. Clearly, the solution plan must involve the production of more timber. However, a valid relaxed plan solution, found using the Metric RPG heuristic, is:

```
0:  (load v1 p1 timber)
1:  (unload v1 p1 timber)
```

### 3.2.3 Helpful Action Distortion

The problems of resource persistence and CRT have an important consequence for EHC search. Not only do they result in relaxed plans with misleadingly short lengths, but the relaxed plans typically contain too few production actions and useless transfer actions. In this situation, production actions often do not appear in the helpful action set, and are therefore not included in EHC search, even in states where they could conveniently be applied. We refer to this problem as *helpful action distortion*. To illustrate how this arises, again with reference to the Settlers domain, consider a state in which there is a unit of timber at location $a$, and the goal is to have a unit of timber at location $b$. A relaxed plan can use the timber to construct a cart and then load *the same* timber onto the cart to transport it. The planner will therefore not consider producing more timber.

## 4. Compiling Producer–Consumer Behaviour into a Mathematical Program

In this section we describe how a mathematical program can be built to characterise the interaction between numeric variables and action choices in the producer–consumer framework.

### 4.1 Constraints for Producer–Consumer Variables and Actions

The definition of producer–consumer variables (Definition 2.4) implies that actions have the useful property that all preconditions on the variables can be derived from the effects of actions, together with the global variable bounds. Specifically, for each action $a$:

- If $a$ produces $c$ units of $v$ and has a precondition requiring $v \leq d$, then $d = ub(v) - c$, where $ub(v)$ is the global upper bound on $v$ after $a$ has been applied. This condition expresses both the effect of $a$ on the upper bound and the precondition on the value of $v$ (since $v \leq ub(v)$).

- If $a$ consumes $c$ units of $v$ then $v$ must satisfy $v \geq lb(v) + c$ before the action is applied. Again, this expression leads to the effect and precondition being tied into one constraint.

If the ordering of actions is relaxed (that is, the causal relations that force them to be ordered are ignored) then the value of $v$ after a series of actions has been applied, $v'$, is given by:

$$v' = v + \sum_{a \in A} C_a . \delta(v, a) \qquad (1)$$

where $C_a$ is a non-negative (count) variable indicating how many times the action $a$ is applied, $v' \in [lb(v), ub(v)]$ and $\delta(v, a)$ is defined as follows:

- If $a$ produces $c$ units of $v$ then $\delta(v, a) = c$;

- If $a$ consumes $c$ units of $v$ then $\delta(v, a) = -c$;

- Otherwise, $\delta(v, a) = 0$.

Note that this equation is linear, since Definitions 2.2 and 2.3 require that $\delta(v, a)$ is constant for any $v$ and $a$.

These equations support the construction of a mathematical program consisting of one variable for each action, $a$, and one variable and flow equation for each producer–consumer variable, $v$. The program is, in fact, a mixed integer program (MIP), because the variables (the action counts, $a$) represent applications of actions, which can only be integral. However, a further relaxation can be exploited to allow the action count variables to take non-integral values, yielding a linear program (LP). The significant potential benefit of doing so is that LPs can be solved far more efficiently than MIPs.

## 4.2 Bounding Action Variables

Within the equation associated with each state variable (i.e. each Equation 1), each action has a corresponding variable denoting how many times it has been applied. In general there is no limit on the number of times an action can be applied. However, numeric decrease effects and propositional delete effects may impose constraints in practice, due to limited availability of resources. For example, if applying an action $a$ increases $v$ at the expense of decreasing $w$, where $w$ is a resource for which there is no producer, then the limit on $C_a$ will be an implicit consequence of the instance of Equation 1 governing the value of $w$. Specifically, $a_i$ can never exceed $w/\delta(w, a)$: the value of $w$ divided by the change $a$ causes in $w$. Moreover, because $w$ is monotonically decreasing and $\delta(w, a)$ is constant for each action $a$, the global upper bound on $C_a$ can be set to $w(I)/\delta(w, a)$, where $w(I)$ is the value of $w$ in the initial state, $I$.

If an action $a$ increases $v$ at the expense of irreversibly deleting some fact $p$, a fact it has as a precondition, then clearly $a$ can only be applied once — it is a one-shot action (Coles, Coles, Fox, & Long, 2009). However, in contrast with the numeric delete effects discussed above, this will not be captured by the producer–consumer constraints which are concerned only with *numeric* change. However, the constraint on the use of $a$ can be captured by setting the upper bound on $C_a$ to 1. Moreover, if a collection of actions $a_0...a_{n-1}$ each depend on a fact $p$ and each irreversibly delete it, we can say that they form a one-shot action set over $p$, and add the constraint:

$$C_{a_0} + C_{a_1} + ... + C_{a_{n-1}} \leq 1 \qquad (2)$$

## 4.3 Assignment Constraints

In general, direct assignments of values to variables cannot be represented directly in constraints following the form of Equation 1. Assignments correspond, effectively, to state-dependent increases or decreases. For instance, an assignment of the value 2 to a variable $v$, in a state where $v = 0$, is equivalent to producing 2 units, but is equivalent to consuming 1 unit when applied in a state where $v = 3$. However, in the producer–consumer equations there is no notion of state, only coefficients on action variables to denote their production or consumption. Therefore, the state variables can only be subject to constant-valued change and the MIP cannot be used to encode general assignment effects without extending it to allow quadratic constraints (that is, constraints involving products of pairs of variables). However, there are some specific conditions under which assignments can be safely modelled within the mathematical program while retaining the linearity of constraints.

One class of assignment effects that can be encoded in the MIP is that in which the actions with assignment effects to a variable, $v = k$, can only be applied in states in which $v = c$ for some known constant, $c$. In this case, the effect can be rewritten as an increase effect on $v$ of $k - c$, making the assignment actions follow the standard pattern for producer/consumer actions. A particular case in which this rewriting is made possible is if the following conditions hold of the set of actions $A$:

1. No action can depend upon or affect $v$ before some condition is satisfied that is achieved by each of the actions in $A$ and only by actions in $A$.

2. Applying any action in $A$ precludes any further assignments to $v$.

These conditions ensure that the set of actions that can assign to $v$ form a one-shot action set and the value of $v$ can be assumed to be 0 prior to application of one of the actions in this set, with each action in the set being rewritten to increase $v$ by its assignment value. This situation is one that arises in encodings in which objects are created by certain actions and those objects have associated metric variables that are initialised on object creation (such as the capacity of a newly created vehicle in the Settlers domain).

## 5. The Linear Programming–Relaxed Planning Graph Heuristic

We have defined producer–consumer behaviour and shown how it supports construction of a MIP in which action ordering is relaxed. A MIP is, in principle, hard to solve (constructing solutions is NP-hard), so as the basis of a heuristic evaluation of states it seems sensible to further relax the integrality constraints on action variables to reduce the problem to a linear program. We will first consider how the LP can be used in the two stages of heuristic evaluation of a state (reachability analysis and relaxed plan extraction) and then we will reconsider the question of whether the relaxation to a linear program is necessary in practice and what compromise there might be between the full MIP and the LP.

### 5.1 Overview

The context for the use of the LP is in a forward state-space search planner. The task for which we intend to use it is the heuristic evaluation of states. Thus, we can assume that we have a state (a complete assignment to the variables that define the problem, both propositional and numeric) and we are interested in estimating the number of actions that will be required to transition to a goal state. The approach we will use is based on the same strategy as is used in MetricFF: first construct a reachability analysis using a layered graph of alternating facts and actions and then extract a relaxed plan. To determine whether an action can be applied in the reachability stage we check propositional preconditions in the usual way and numeric preconditions are checked by determining whether some values in the reachable ranges recorded for the metric variables will satisfy the condition. The reachable ranges are calculated using the LP we have described, as we explain below.

Extraction of the relaxed plan involves determining which actions support required conditions (both goals and preconditions of selected actions), and where the conditions involve numeric variables we use the LP to decide which actions will be required and how many of them will be used.

### 5.2 Using the LP during Graph Expansion

The graph expansion phase in calculation of the Metric RPG heuristic can be seen as a layer-by-layer relaxed propositional reachability analysis, synchronised with a relaxed numeric bounds analysis. Definition 3.1 shows that numeric variable bounds appear in the graph expansion algorithm in two places. First, at $F(i)$ they are used to determine which actions can appear in $A(i+1)$, the next action layer (those whose preconditions are within the reachable range). Then, the actions deemed applicable are used to update the variable bounds for the subsequent fact layer, $F(i+1)$.

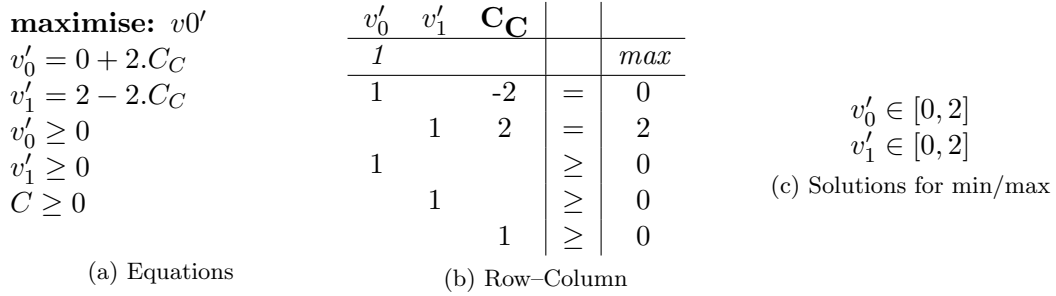

Figure 2: LP to maximise the value of $v_0$ in layer $F(i+2)$ of Figure 1 (treating layer $F_i$ as the initial state for this construction)

Due to the relaxed nature of the way numeric values are considered, the metric RPG tends to produce highly optimistic bounds on the numeric values. Returning to Figure 1, we can see that, in effect, the action $C$ converts two units of $v_1$ into two units of $v_0$ and, initially, just two units of $v_1$ are present. Hence, at most, we could hope to produce two of $v_1$, but in $FV(i+2)$, the upper bound on $v_1$ is already 4. Ignoring the consumption effect of $C$ makes it possible to produce arbitrary amounts of $v_0$: $C$ is applicable if the upper bound on $v_1$ is greater than or equal to two, which in reality is only true so long as $C$ has not yet been applied.

A more accurate estimate of the variable bounds in layer $FV(i)$ can be found using the LP encoding described in Section 4.1. The model is parameterised as follows:

- Only those action variables corresponding to actions in $A(i)$ are used, which ensures that only reachable actions are considered in computing resource bounds. Any relevant one-shot action constraints are included. The absence of any restriction on the number of action applications contrasts with the constraint used in Metric FF that each action may only be applied once per action layer. In practice to prevent the LP variables becoming unbounded, we set a finite, but large maximum value for action variables.

- The initial value of each variable $v$ is set from the state, $S$, being evaluated.

- The post-value of a variable, $v'$, is $FV(i)[v]$, the range of values it could reach by $FV(i)$. As always, $v' \in [lb(v), ub(v)]$.

Substituting these parameters into the producer–consumer equation (Equation 1) yields:

$$FV(i)[v] = S[v] + \sum_{a \in A(i)} C_a . \delta(v, a) \qquad (3)$$

With this model, we can then use an LP solver to find upper and lower bounds on each $FV(i)[v]$, by setting the objective function accordingly.

Returning to our example, consider finding the upper bounds on the variables in $F(i+2)$ of Figure 1 starting from a state corresponding to the one given as $F_i$ in that figure (thus, we are considering the constraints in layer 2 following the state we are treating as our starting point). The corresponding LP is shown in Figure 2, where the primed variables are the ones we use to represent the values of the numeric variables in the layer of interest (i.e. in this case the layer 2 ahead of the state being evaluated). Maximising $v_0'$, i.e. using the upper bound of $FV(i)[v_0]$, yields the result 2: no greater value is possible, since setting $C$ to a value greater than 1 (and thus producing more $v_0$) would lead to a violation of the constraint $v1' \geq 0$. The ranges of the variables computed, using the LP four times (minimising and maximising each of the two variables), are shown in Figure 2c. As can be seen, these improve over the bounds calculated in the same situation by MetricFF (Figure 1), which are $v_0' \in [0, 4]$ and $v_1' \in [-2, 2]$ respectively.

### 5.2.1 NOTES ON LP EFFICIENCY

As the LP has to be solved up to twice per variable per layer when expanding the planning graph, it is important that steps are taken to minimise the computational cost. We reduce the costs using a combination of techniques, some that avoid needing to solve the LP when computing the bound on a given variable, and others which reduce the cost of solving the LP itself.

1. First, as a consequence of the termination criteria for RPG expansion (Section 3.1.1), there is no need to compute the upper (lower) bound on a given variable if its current value is large enough (small enough) to satisfy all the preconditions and goals in which it appears. In this case, we can avoid having to solve the LP to determine the variable bound, and instead can re-use the bound computed at the previous layer, without affecting the behaviour of the heuristic.

2. If a variable never appears in a numeric precondition or a goal, it can be entirely excluded from the LP.

3. The bounds on variables change monotonically as additional layers are added to the planning graph. Therefore, when computing the new upper (lower) bound on a variable $v$ we can temporarily add to the LP the constraints corresponding to the bounds computed at the preceding layer (each is added separately as the variable is first minimised and then maximised). By doing so, we refuse to admit a tighter variable bound than in the previous layer.

4. Finally, If no actions with an increase (decrease) effect on a variable have yet been added to the LP, we do not need to compute the upper (lower) bound on the variable, as there is no effect by which the bound on the variable can be increased (decreased) beyond the value in the state being evaluated.

## 5.3 Basic Use of the LP during Solution Extraction

We now consider how the LP can be used to give guidance in action selection during relaxed plan extraction. First, we observe that the LP is not directly affected by propositions, and hence cannot be used to find which actions are able to achieve a given fact. Thus, we concern ourselves with how the LP can be used to identify which actions to use to attain numeric subgoals — either the top-level numeric goals, or the numeric preconditions of actions chosen during solution extraction.

---

**Algorithm 3**: Adding a weighted action to a relaxed plan

---

**Data**: $R$ — a metric RPG, $a$ — an action to include in the relaxed plan, $q$ — the subgoal queue, $w$ — a weight

**1 foreach** *propositional precondition pre of a* **do**
**2**     $l \leftarrow$ layer at which *pre* first appears;
**3**     **if** $l > 0$ **then**
**4**        **if** $\exists (pre, k) \in q[l].prop$ **then**
**5**           $\lfloor$ **if** $k < w$ **then** $k \leftarrow w$;
**6**        **else** insert $(pre, w)$ into $q[l].prop$;

**7 foreach** *numeric precondition pre of a* **do**
**8**     $l \leftarrow$ layer at which *pre* first appears;
**9**     **if** $l > 0$ **then**
**10**    **if** $\exists (\{pre\}, k) \in q[l].num$ **then**
**11**        $\lfloor$ **if** $k < w$ **then** $k \leftarrow w$;
**12**    **else** insert $(\{pre\}, w)$ into $q[l].num$;

---

The first key difference when using the LP during the relaxed plan extraction concerns the choice of actions to achieve numeric preconditions. In the original metric RPG heuristic, a given numeric precondition (e.g. $x \geq c$) in fact layer $i + 1$ was regressed through all the beneficial numeric effects at layer $i$, giving a residual numeric precondition (e.g. $x \geq c'$) to then be achieved in fact layer $i$. (This process is shown in lines 28 to 41 in Algorithm 2). In the LP-RPG case, shown in Algorithm 4, a numeric precondition at layer $l$ is (temporarily) added to the LP generated for layer $l$ (as a constraint (line 22). To find the actions to use to achieve this, the LP is solved (line 23), with the objective being to minimise a weighted sum across the action variables (one possible weighting scheme that will suffice for this purpose is to minimise the sum of the action variables, though we return to the question of appropriate weighting schemes later in the paper). Finally, the actions whose corresponding variables are non-zero (line 24) are added to the relaxed plan (lines 25 to 34).

Second, we must accommodate the fact that the LP, being a relaxation of an underlying MIP (in which the action variables are integers), may be solved by applying actions a non-integral number of times. As a simple example, if there are several actions that increment a given variable, and any of them alone would suffice to achieve a goal value for the variable, then a valid optimal solution to the LP is any for which the sum of the variables corresponding to these actions is 1. If every action that had a non-zero action count variable in the solution were considered 'applied' then the relaxed plan length could greatly over-estimate the required number of actions. To mitigate this problem, each subgoal (e.g. $x \geq c$) arising during solution extraction is associated with a weight, and these weights, along with the values given to action variables in the LP, are used to update the relaxed plan length. The weights are manipulated throughout Algorithms 3 and 4, and their use can be summarised as follows:

- Initially, each goal fact is added to the subgoal queue with associated weight 1, i.e. each has to be achieved, entirely. Also note that, in contrast with Algorithm 1, the

---

**Algorithm 4**: Relaxed plan extraction with LP

---

**Data**: $R$ - a metric RPG; $PG$ - propositional goals;
      $NG$ - numeric goals

**Result**: $ha$ - helpful actions, $h$ - A heuristic value

**1**   $ha \leftarrow \emptyset$, $h \leftarrow 0$;

**2**   $q \leftarrow$ deepest-first priority queue of goal layers;

**3**   **foreach** $p \in PG$ **do**

**4**      $l \leftarrow$ layer at which $p$ first appears;

**5**      insert $(p, 1)$ into $q[l].prop$;

**6**   **if** $|NG| > 1$ **then**

**7**      $l \leftarrow$ final layer of $R$;

**8**      insert $(NG, 1)$ into $q[l].num$;

**9**   **else**

**10**      $f \leftarrow$ the fact in $NG$;

**11**      $l \leftarrow$ layer at which $f$ first holds;

**12**      insert $(\{f\}, 1)$ into $q[l].num$;

**13**   **while** $q$ *not empty* **do**

**14**      $(l, \langle prop, num \rangle) \leftarrow pop(q)$;

**15**      **foreach** $(p, w) \in prop$ **do**

**16**          $h \leftarrow h + w$;

**17**          $a \leftarrow$ achiever for $p$;

**18**          **if** $a$ *in action layer 1* **then** add $a$ to $ha$;

**19**          call Algorithm 3 with $R, a, q, w$;

**20**          $prop \leftarrow prop \setminus$ add effects of $a$;

**21**      **foreach** $(G, w) \in num$ **do**

**22**          LP' $\leftarrow$ LP($l$) + the constraint(s) $G$;

**23**          solve LP', minimising weighted action sum;

**24**          $av \leftarrow \{action\ variable\ (\mathrm{a} = \mathrm{c}) \in$ LP' $\mid c \neq 0\}$;

**25**          **foreach** $a \in av$ **do**

**26**             $h \leftarrow h + w.c$;

**27**             **if** $a$ *is in layer 1* **then** add $a$ to $ha$;

**28**             **foreach** *propositional precondition pre of $a$* **do**

**29**                $l \leftarrow$ layer at which *pre* first appears;

**30**                $c \leftarrow \min[c, 1]$;

**31**                **if** $\exists (pre, k) \in q[l].prop$ **then**

**32**                    **if** $k < w.c$ **then** $k \leftarrow w.c$;

**33**                    remove $(pre, k)$ from $q[l].prop$;

**34**                **else** insert $(pre, w.c)$ into $q[l].prop$;

---

subgoal queue here records the layer at which the goal is introduced as well as its associated weight.

- An action $a$ can be chosen to be applied $C_a$ times to achieve a queued propositional/numeric sub-goal $g$, if either:

  - it is chosen to be applied once (i.e. $C_a = 1$) as the achiever for some propositional subgoal $g$, with associated weight $w$;

- the action was applied (given a non-zero value for $C_a$) when solving the $LP$ to achieve some numeric subgoal $g$, with associated weight $w$.

- In both of these cases, the relaxed plan length is incremented by $C_a.w$.

- The weight given to the preconditions of $a$ is then $w' = w.min[C_a, 1]$. The weights of these preconditions are used to update the weight attached to achieving the corresponding sub-goals at earlier layers:

  - If a propositional precondition $p$ of $a$ is already a recorded subgoal at an earlier layer, with some weight $k$, then its weight is updated to be $max[k, w']$. Otherwise, $p$ is added as a subgoal to the RPG, to be satisfied in the first layer at which it appears, with weight $w'$.

  - In the case of $a$ being added to support a propositional sub-goal: if a *numeric* precondition $p$ of $a$ is already a recorded subgoal at an earlier layer, with some weight $k$, then its weight is updated to be $max[k, w']$; otherwise, $p$ is added as a subgoal to the RPG, to be satisfied in the first layer at which it appears, with weight $w'$.

## 5.4 Consequences of the Use of the LP during Solution Extraction

Use of the LP to aid in the identification and selection of actions to support achievement of numeric goals and subgoals in the extraction of a relaxed solution can lead to some important consequences on the heuristic guidance offered by the relaxed solution. We have already noted the problem that non-integral fragments of actions might be combined to achieve numeric effects and indicated that this can be managed by handling fractional preconditions and fractional action costs. However, there are other potential problems as we now discuss.

### 5.4.1 Partially Applied Helpful Actions

Consider a situation in which there are precisely five possible ways to achieve a particular numeric goal, and each of these uses three actions. A simple example of this is a small problem in the Settlers domain, where there are five carts and a unit of timber available at location $A$: the goal of having one unit of timber at location $B$ can be achieved by loading timber onto any of the carts at $A$, moving it from $A$ to $B$, and then unloading it at $B$. If the metric RPG heuristic were used to achieve this goal, three actions would be used. Working backwards from the goal, the selected actions would be:

- in action layer three, an unload action (from cart $c$) to increase the amount of timber at $B$;

- in action layer two, an action to move a cart $c$ from $A$ to $B$;

- in action layer one, an action to load a unit of timber at $A$ onto the cart $c$.

When solving the LP to achieve the same goal (ignoring propositional preconditions), with the objective of minimising the sum of the action variables, the solution returned will have an objective value of 2. If we denote the relevant load/unload action variable pair for cart $i$ as $(l_i, u_i)$, the pool of solutions that could be returned is any satisfying:

$$\underset{i \in [1..5]}{\forall} l_i = u_i \qquad (\sum_{i \in [1..5]} (l_i + u_i)) = 2$$

Then, for any non-zero variable $u_i = k$ the relevant action to move cart $i$ from $A$ to $B$ will also be added to the relaxed plan, with weight $k$ (line 19, Algorithm 4).

For the purposes of providing a contribution to the relaxed plan length, it is unimportant which of these solutions is returned: the sum of the action variables in each is 2, and the sum of the move actions added is 1, giving a total relaxed plan length of 3. However, as discussed in Section 3.1.2, the relaxed plan is also used to determine a set of *helpful actions*: those with an effect in common with the actions in the relaxed plan that were chosen from action layer one. In this example, action layer one consists of the action (or actions) to load a unit of timber onto a cart at $A$. In the original metric RPG case, exactly one action would be used. However, using the LP up to five actions could be (fractionally) used. The consequence of this is that within the pool of LP solutions that could be returned, some will lead to search having a much greater branching factor, in this case up to a factor of five greater.

The source of this problem is the relaxation of integrality constraints on the action variables and the extent to which it affects search depends on the precise solution returned by the LP solver: different solvers may have a greater or lesser tendency to return solutions in which the action variables are assigned non-integral values. An extreme response to this problem would be to revert to the MIP and require all action variables to be integral rather than real-valued. Alternatively, focussing on the issue identified here, one could require only the action variables corresponding to actions from action layer one to be integers. In either case, the result is a mixed integer programming problem, but since the cost of MIP-solving is exponential in the number of integer variables, the difference between the variants can be significant. This change only need be made at the point where we switch to using the LP for solution extraction rather than graph expansion as, prior to this, the assignments to the action variables are unimportant (only the value of the objective function is used). Of course, the price we pay is potentially very significant, because MIP-solving is NP-hard, while LP-solving is polynomial. However, in exchange for this shift in complexity, in the example given above, we are left with only a single helpful action, as it is no longer possible to fractionally load timber onto a cart: one cart must be chosen. The extent to which these two possible integer-modifications affect search performance will be considered later in the evaluation.

### 5.4.2 PREFERRING EARLIER ACTIONS

Within the Metric RPG heuristic there is an explicit preference for using actions that appear earlier in the relaxed planning graph. As shown in Algorithm 1, when a fact is needed, either as a goal or to satisfy a precondition of an action chosen for insertion into the relaxed plan, it is queued as a sub-goal to be satisfied at the *first* fact layer in which it appeared. Then, when an action is chosen to support that fact, it will be amongst the earliest possible achievers. The intuition behind this preference for earlier actions is based on the observation that, the later an action appears in the relaxed planning graph, the greater the number of actions that need to be added to the relaxed plan to support its preconditions. Therefore, preferring earlier actions usually leads to shorter relaxed plans and hence to closer approximations of the optimal relaxed plan length.

Within the LP, if the objective is set to minimise the sum of the action variables (i.e. use as few actions as possible), then there is no distinction between actions that appear earlier in the RPG and those that appear later. Recalling that the LP disregards the propositional preconditions of actions, failing to take into account when an action is first added to the RPG can lead to LP-based relaxed plan extraction generating very poor quality solutions and, consequently, very poor search guidance.

To address this, the pressure generated within the LP by the objective function has to be tuned to prefer actions that need fewer supporting actions in the relaxed plan. This is achieved by forcing the LP to favour actions that appear in the RPG earlier. We encode a preference for earlier actions with the weighting scheme for the action variables in the objective function. Actions appearing earlier are given smaller weights than those that appear later. We propose (and, later, will evaluate) two ways of achieving this. The first, and simpler, is to use a geometric series to dictate the coefficient given to an action variable $a$ based on the layer $l$ in which it first appears. In this case, the objective coefficient on $a$ is:

$$k^l \qquad k \in \Re \wedge k > 1.$$

The value of $k$ controls the extent to which earlier actions are preferred, and can be interpreted as treating $k.n$ actions selected from layer $l$ as exactly as good as selecting $n$ actions from layer $l+1$ (so anything less than $k.n$ actions from layer $l$ will be preferable to selecting $n$ actions in layer $l+1$). Throughout the remainder of the paper, we refer to this scheme as layer-weighting with value $k$.

The second option is to record, as a cost for each action, an estimate of the number of actions needed to support its propositional preconditions, and use this as its weight in the objective function. This can be achieved by using the RPG cost propagation algorithm from SAPA (Do & Kambhampati, 2003). To achieve this, as the planning graph is expanded, costs for each fact and action are recorded and updated at each layer. Initially, for each fact $p$ in the state being evaluated, its cost at fact layer zero, $cost(p, 0)$, is zero. (For each fact $p$ not true at time zero, $cost(p, 0) = \infty$.) These fact costs are then used to derive action costs, using rules akin to those used by $h^{add}/h^{max}$ (Bonet & Geffner, 2001). The cost of an action $a$ at layer $t$, $cost(a, t)$ is defined according to one of either:

$$cost^{max}(a, t) = \max_{p \in pre(a)} cost(p, t-1)$$

$$cost^{sum}(a, t) = \sum_{p \in pre(a)} cost(p, t-1).$$

These action costs, in turn, are used to update the costs of each proposition in the subsequent fact layer, with the cost of each fact being reduced if there is now a cheaper way to achieve it. For an action $a$ in layer $t$, it can potentially reduce the cost of each of the propositions $p$ that it adds:

$$
\begin{aligned}
cost(p, t) \quad &= cost(a, t) + 1 \quad \text{iff } (cost(a, t) + 1) < cost(p, t-1) \\
&= cost(p, t-1) \quad \text{otherwise}
\end{aligned}
$$

As the planning graph is expanded, this process of alternating action cost estimation, and fact cost estimation, is used to propagate cost information through the RPG. When setting the objective of the LP, we can use the cost of an action, $cost(a, t)$, as the coefficient of the action variable corresponding to $a$. Using the cost propagation as described, the costs are derived solely on the basis of propositional preconditions and effects. Therefore, $cost(a, t)$ is an estimate of the number of actions needed to support the preconditions of $a$. For our purposes, this is desirable: the LP will, itself, add actions to support numeric preconditions, so an estimate of the number of actions needed to support the propositional preconditions of an action is a measure of the cost impact upon the relaxed plan due to an action being selected.

## 6. Adding Propositions to the LP

Benton et al. (2005) explore the idea of using an LP to guide search in propositional planning problems in the context of over-subscription planning. In that work the LP is used to determine which goal subset to achieve, to gain maximum utility. Whilst successful in achieving that aim, the authors observe that the use of the LP as a heuristic to guide search is very expensive, indeed too expensive to be feasible. With the propositions encoded in the LP in the way they propose, the task of solving the LP becomes equivalent to solving the entire planning problem with relaxed action ordering and non-integer action variables. In this section, we reconsider the inclusion of propositions in the LP, considering the spectrum of possibilities between including no propositions and a way to include all propositions.

Even though the focus of this work is on numeric problems, including some propositions in the LP might still be of interest. For instance, supporting a propositional goal might require the consumption of numeric resources. In the worst case, one could compile a problem so that all the numeric goals become preconditions of an action that achieves a dummy propositional fact `goal`, and modify the problem so that the only goal is `goal`. Since there are no numeric goals in this modified problem, the LP will then only be used to solve the individual preconditions of the dummy action when it is (inevitably) chosen, rather than requiring the goals to be satisfied in conjunction, as described in Section 7.1. Since this dummy-goal model is merely a reformulation of the original problem, the same information should theoretically be accessible to be conveyed to the LP. More generally, we can hope to identify intermediate landmark propositions, as well as final goals, that could usefully be encoded in the LP.

### 6.1 Adding Propositional Goals to the LP

Although the LP we describe in Section 4 does not contain specific reference to propositions, and hence propositional goals, we can formulate constraints that act as a proxy for them, by considering which actions achieve them. We do not need to introduce additional variables. Instead, we add constraints to ensure that at least one achiever is chosen for each propositional goal. For each goal fact $g$ that is not true in the state being evaluated, then for the list of actions $[a_0..a_{n-1}]$ that achieve $g$, we can add as a constraint to the LP for the most recent layer in the planning graph:

$$a_0 + ... + a_{n-1} \geq 1.$$

That is, at least one achieving action must be used or, more specifically, given actions can be partially applied, a total of at least one achieving action must be used. An LP containing these constraints can be used to augment the positive termination criteria for graph expansion, detailed in Section 3.1.2. We then terminate at the first fact layer $i$ where:

1. All goal propositions $F^\star$ are in $FP(i)$ (as before);

2. All goal numeric expressions $N^\star$ are satisfied (individually) by $FV(i)$ (as before);

3. The LP used to compute the numeric bounds for layer $F(i)$ is still solvable when all the constraints for propositional goals are added.

Use of the goal-checking LP has two key consequences. First, if the actions up to layer $F(i)$ cannot be used to satisfy the goals whilst respecting the other numeric constraints in the LP, additional layers are added to the planning graph until the necessary actions have appeared (or the termination criterion is reached). Thus, in reasoning about resource persistence (Section 3.2.1), the heuristic is now better able to recognise cases where, although the propositional goals might appear to be individually reachable, either additional production is needed to meet them collectively, alternative actions need to be used, or the state is a dead-end. Second, the solution to the LP used to confirm point (3) above, is used to indicate which actions to add to the relaxed plan to achieve the propositional goals. The propositional preconditions of these actions will be satisfied in the usual way (line 28 of Algorithm 4).

By requiring only that the sum of the action variables selected to achieve the goals is at least one, and allowing such variables to be real-valued, the LP could, in theory, provide weaker guidance than the RPG. This is a similar issue to that noted in Section 5.4.1 when considering helpful actions, and could be ameliorated in a similar manner, namely by making the goal-achieving action variables integral. We will return to this issue in the evaluation, considering whether or not this benefits search.

## 6.2 Using Landmarks in the LP

A landmark fact (Hoffmann, Porteous, & Sebastia, 2004) is a propositional fact that must be true at some point in *every* solution plan to a given planning problem. The first work on landmarks (Porteous, Sebastia, & Hoffmann, 2001) proposed a method for extracting a subset of the landmarks from a planning problem based on regressing from the goals using the delete relaxation of FF. Since introduction of the idea in 2001 (Porteous et al., 2001), landmarks have have come to play an increasingly important role in planning. Recent development of new techniques for extracting landmarks (Richter, Helmert, & Westphahl, 2008; Zhu & Givan, 2003) and the development of heuristics based on different relaxations (Richter & Westphal, 2010; Domshlak, Katz, & Lefler, 2010; Helmert & Domshlak, 2009; Karpas & Domshlak, 2009) have allowed the planning community to exploit landmarks more successfully.

The relaxed plan extraction phase of the LP-RPG heuristic relaxes action ordering and propositional preconditions and effects and might benefit substantially from delete-relaxation landmarks. The use of landmark facts in the LP offers a further mechanism by which to more tightly couple the LP and the RPG, allowing increased information sharing between the propositional and numeric components of the heuristic. If we know that a landmark fact must occur in any solution plan, and it has not yet appeared on the path to a state being evaluated, we can add constraints representing this to the LP, just as we did for propositional goals. That is, that the sum of the action variables $[a_0..a_{n-1}]$ achieving a given landmark must be greater than or equal to 1. As with propositional goals, this constraint introduces the need to provide numeric support for the action(s) chosen to support the landmark. Goals are a special case of landmarks, but an important feature of goals is that, even if they have been achieved on the path to the current state, if they are not true in the current state then they must be reachieved. Constraints can be added to the LP to ensure this. To reflect landmarks that have been achieved on the path to the current state, the state is modified to record them and the record updated as new landmarks are seen. This approach is similar to LAMA (Richter & Westphal, 2010).

A set of *disjunctive landmarks* is a set of propositional facts, any one of which must be true in any solution to a planning problem. The extraction of disjunctive landmarks has been considered (Gregory, Cresswell, Long, & Porteous, 2004) but it is even more difficult than is the case with conjunctive landmarks to exploit them in planning systems. The knowledge that a certain fact must be true can allow the planner to infer that certain actions that must be present in solution plans, which can inform heuristics. However, disjunctive landmarks are less informative. Disjunctive landmarks often arise from problem symmetry. For example, we might know that in order to deliver a package from one place to another it will have to be loaded in to a truck, but not which truck. A disjunctive landmark in which the package is in some truck can be generated: even if we do not know which truck to use, we know that one of the disjunctive landmarks must hold. Thus, in the context of numeric resources, some truck must be fueled (assuming they all start empty), which might entail other additional costs. It is therefore of interest to take account of disjunctive landmarks in numeric reasoning.

We are able to make use of disjunctive landmarks in the LP to further constrain the problem, ensuring that support is given to at least one fact within each (unreached) disjunctive landmark. When dealing with standard conjunctive landmarks, the constraint is that a sum of at least one achiever must be added for each propositional landmark. For disjunctive landmarks, however, the constraint is slightly different. A disjunctive landmark set $L$ is satisfied if any of its constituent landmarks are satisfied. That is, if we apply any of the actions:

$$achieves(L) = \{a \mid \textit{eff}^+(a) \cap L \neq \emptyset\}.$$

To encode the disjunctive landmarks in the LP (assuming it has not yet been met) there are two possibilities. The first is to add a binary variable $sf$ for each fact $f \in L$, with constraints such that $sf$ can only take the value 1 if there is at least a total of one action adding $f$, and that at least one such variable $sf$ has to take the value 1. That is, at least one of the disjunctive landmarks has to be fully met. An alternative, potentially cheaper, approach is to add a constraint that $(\sum achieves(L)) \geq 1$. This allows the disjunctive landmark to be considered satisfied if the sum across the action variables supporting any of its constituent facts is at least 1. This is somewhat weaker than the constraint for individual, non-disjunctive landmarks, as it does not guarantee that there is support of at least 1 for any *individual* constituent fact. For instance, a two-fact disjunctive landmark is 'satisfied' if the support for each constituent fact is 0.5. We considered both of these approaches and found that there is a negligible difference in performance (time taken and nodes expanded) between the two encodings so no real saving is achieved by using the relaxed approach.

## 6.3 Managing Propositional Preconditions and Effects

So far, we have considered propositions that must be achieved in a planning problem due to goals and landmarks. However, there is a second class of propositions: those that, given the values assigned to the action variables in the LP, must also have supporting actions added to the solution relaxed plan.

To extend the LP to capture propositional preconditions and effects, we first introduce a binary variable $f$ (an integer whose value is 0 or 1) for each fact $f$ that is not true in the state being evaluated. This is then involved in two constraints. First, for the actions $[a_0^+...a_{n-1}^+]$ that add $f$:

$$a_0^+ + ... + a_{n-1}^+ \geq f.$$

In the case that the proposition corresponding to $f$ is a goal, $f = 1$ and hence one or more of the achieving actions must have a positive value (since the constraint is expressed using continuous variables, the actions might only be partially applied in the relaxation). Then, for the actions $[a_0^p..a_{m-1}^p]$ that have $f$ as a precondition:

$$N.f \geq a_0^p + ... + a_{m-1}^p$$

where we use $N$ to denote a (sufficiently) large number. This constraint ensures that if at least one of the actions depending on the proposition corresponding to $f$ is used in a relaxed plan, then $f$ must be positive (that is, the corresponding proposition is required to be true within the relaxation). The use of $N$ is to ensure that $f = 1$ is sufficient to satisfy the preconditions of many actions. This pair of constraints is effectively a conditional version of the constraint to meet a propositional goal, described in Section 6.1. In cases where the proposition is neither a goal nor a landmark, these constraints serve to enforce that at least one action that adds $f$ is chosen (has a positive value) in the LP if any action requiring $f$ is chosen even partially.[3]

## 6.4 Recognising Propositional Resources

Finally, we consider one other case where it is potentially useful to model propositions in an equivalent numeric form. In PDDL, finite domain integer resources can be modelled in two ways: as numeric variables, or as a set of propositions. Consider the following two formulations of the `fell-timber` action from the Settlers Domain (for simplicity the effect on the metric tracking variable `labour` is omitted):

```
(:action fell-timber
  :parameters (?p - place)
  :precondition (has-cabin ?p)
```

---

3. It is tempting to consider replacing this constraint, with its slightly troublesome $N$, with constraints of the form $f \geq a_i^p$ for each $i$. Unfortunately, this is not appropriate because an action variable, $a_i^p$, can be greater than 1 (due to multiple applications of the action) and yet $f = 1$ is sufficient to satisfy the precondition of all the action applications.

```
    :effect (increase (available timber ?p) 1)
)

(:action fell-timber
   :parameters (?p - place ?n0 ?n1 - value)
   :precondition (and (has-cabin ?p)
                      (timber ?p ?n0)
                      (less-than ?n0 ?n1))
   :effect (and (not (timber ?p ?n0))
                (timber ?p ?n1))
)
```

Each of these representations models the same situation, but each uses a different mechanism to do so. The first uses numeric variables, while the second uses propositions. When using either the numeric or propositional formulations with the MetricFF heuristic, there is little practical difference in the guidance given. In the numeric case, once the fell-timber action has been applied, the upper bound on the amount of timber at place $p$ is increased. This means that any action consuming this amount of timber can be executed at subsequent layers, regardless of how many other actions using the resource have also been applied. In the propositional case, the delete effect on timber (i.e. that deleting the fact that there previously was some) is also relaxed so, again, any number of actions requiring this amount of timber can be applied.

Turning our attention to the LP-RPG heuristic, however, we can observe that although the RPG part of the heuristic exhibits the same weakness as the propositional case, the different relaxation used in the LP for numeric reasoning means that the consumption of timber would not be disregarded. The LP relaxes action ordering, rather than delete effects (or production/consumption effects), so if the resource is modelled numerically, this interaction can be captured and accounted for. It is therefore in our interests when using LP-RPG to convert resources modelled propositionally into a numeric formulation, so that these can be reasoned with in the LP, rather than in the RPG.

Although the formulation of resources in the above example is an instance of a common idiom used to capture numeric resources in a propositional encoding, there are situations in which it is more natural to model resources propositionally from the outset. This is often the case with binary resources: resources that are either present or not. Such resources are, of course, a special case of the more general resource model described above. Consider the propositional and numeric counterparts of an action to switch on a water pump:

```
(:action activate
   :parameters (?p - pump)
   :precondition (off ?p)
   :effect (and (not (off ?p))
                (on ?p))
)

(:action activate
   :parameters (?p - pump)
```

```
:precondition (<= (pumping ?p) 0)
:effect (increase (pumping ?p) 1)
)
```

Corresponding actions can similarly be created to switch the pump off (the fact `(on ?p)` is deleted and `(off ?p)` added, or equivalently a unit of `(pumping ?p)` is consumed). In many senses, the most natural formulation of this action is the first, using propositions. This is the way most binary resources are encoded in benchmark domains. However, the second formulation is equivalent (assuming the value of `(pumping ?p)` in the initial state is 1 or 0). If there is no interaction between a propositional resource and the other resources identified in the planning problem, there is little motivation to add it to the LP, since no numeric support is required. In the case where a binary resource has an impact on another numeric variable it is, as we shall see, most efficient to model both as numeric resources. Suppose we have some water pumps that can control the flow of water. Two ways to model this in PDDL are shown below:

```
(:action activate
:parameters (?p - pump)
:precondition (off ?p)
:effect      (and (increase (water-flow) 1)
                  (not (off ?p))
                  (on ?p))
)

(:action activate
  :parameters (?p - pump)
  :precondition (<= (pumping ?p) 0)
  :effect (and (increase (pumping ?p) 1)
               (increase (water-flow) 1))
)
```

The first of these two actions switches on a pump (a binary, propositional resource) and produces a unit of `(water-flow)`. If other actions in the domain have preconditions over the water flow, such as an action to run a water wheel with a precondition ($\geq$ `(water-flow)` 3) then there is an interaction between the propositional and numeric variables of the problem. If we use the first model of the action, the RPG will capture the propositional part of the action (whether the pump is on or off) and the LP will encode only the numeric part of the action. Since the RPG relaxes delete effects it will not represent the fact that `(off ?p)` is no longer true and, hence, will not prevent the pump from being switched on many times. In the LP built using the first formulation, the action *activate* does not consume any numeric resources, so it can be used arbitrarily often to increase the water flow — the fact that switching off is necessary to achieve this increase is ignored. Thus, mixing the propositions and numeric resources in the action degrades the information available from the LP.

Using the second formulation, the state of the pump appears in the LP as a variable, and if we use *activate* and *deactivate* to denote the actions for activating and deactivating a pump, the constraints on the *pumping'* variable are:

$$pumping' = init + activate - deactivate$$
$$pumping' \geq 0$$
$$pumping' \leq 1$$

It is now clear that the *activate* action can only be applied once: if it is applied again then *deactivate* must be applied, with the corresponding effect on `water-flow`, in order to satisfy the last of the above constraints. This provides useful guidance, as it indicates that the water flow cannot reach 3 units using only these actions: actions to control more pumps, or other means of increasing flow, need to be added to the LP, through further expansion of the planning graph. If no means to attain sufficient flow can be found, then a dead-end has been discovered that would otherwise have wasted search effort.

Static-analysis techniques capable of identifying propositional resources have been developed (such as the TIM system described in Long & Fox, 2000). These can be used in a preprocessing stage to recognise propositional resources in planning domains and translate them into equivalent numeric resources. We use this translation approach for all recognised resources, with the resulting numeric preconditions and effects being included in the LP in the same way as other numeric variables. In doing so, the LP order-relaxation rather than the RPG delete-relaxation is used to compute heuristic values, preventing impossible reuse of the same resource in cases such as that described.

## 7. Extending the Scope of Numeric Reasoning in the LP

In Section 4 we discussed an LP encoding that captures the producer–consumer behaviour of actions, as used in the first version of LPRPG (Coles et al., 2008). In this section we discuss how this encoding can be enhanced, with the use of further numeric information representing the structure of the planning problem, to improve the guidance that the resulting heuristic can provide to the planner. We address two key issues here: ensuring that conjunctions over numeric goals can be satisfied, and considering the issue of fractionally applied actions in the LP.

### 7.1 Checking Numeric Goals alongside Propositional Goals

In Section 6.1 we noted that we can constrain the LP so that it finds actions to achieve propositional goals. We can extend this further, to capture numeric goals, $N^\star$, adding each numeric goal directly to the LP as a constraint. As with propositional goals, this has clear advantages in terms of resource persistence (Section 3.2.1), by insisting all goals are simultaneously achievable, rather than just individually achievable. Additionally, though, it raises the expressive power of the numeric goals we are able to handle to anything that can be expressed in Linear Normal Form (LNF) — any LNF formula can be added to the LP as a constraint.

### 7.2 Catalytic Resources

So far we have only considered numeric variables that conform to producer–consumer behaviour. However, there is another related class of variables that can also be expressed in the LP in a similar way to producer–consumer variables. These are variables that represent resources that must be *present* in order for an action to be applied, but are not then consumed. The same resources can be used to support many actions[4]. An example of such a resource is a catalyst in a chemical reaction. A catalyst must be either created through reactions, or bought, but once present, it enables other reactions, or allows them to take place more quickly. For these catalytic reactions, the resource must be present, but is not consumed, though it could be that other non-catalytic reactions may consume the resource. Another example one might consider is the building of a unit in a plant, to support some process. The unit must be there in order for the process to occur, but once built it can be used many times to enable other actions without necessitating its destruction. Often in planning problems, the presence of such structures is represented by propositions, but this need not be the case. If many indistinguishable processing units are present, or several units of a catalyst are needed, it often makes more sense to represent these numerically[5].

To extend the LP-RPG heuristic to provide guidance in such problems, where some actions require $v \geq c$ but do not affect the value of $v$, we encounter the difficulty that the LP encodes no notion of time: ordering of actions is relaxed, so it is impossible to ascertain the value of $v$ at a specific time in order to determine whether the (catalyst) precondition is satisfied or not. We therefore add additional constraints to the LP to determine the upper and lower bounds on $v$ obtained for the most optimistic or pessimistic possible ordering of the actions whose variables are non-zero. To find an optimistic upper and lower bound on v, $v \uparrow$ and $v \downarrow$ respectively, we add the constraints:

$$v \uparrow = v + \sum_{\mathsf{a} \in A} a \cdot \max(\delta(v, \mathsf{a}), 0) \qquad (4)$$

$$v \downarrow = v + \sum_{\mathsf{a} \in A} a \cdot \min(\delta(v, \mathsf{a}), 0) \qquad (5)$$

The upper bound is equivalent to ordering all production actions before all consumption actions. For the lower bound, this is reversed, being equivalent to all consumption actions being ordered before all production actions. The bounds are not the same as those computed for the possible values of resource variables in the reachability graph, because here we are considering only the actions that are actually selected for execution in the relaxed plan, rather than those that could possibly be applied. As can be seen, neither requires an explicit notion of time.

Using these bounds, if an action $a$ with a precondition $v \geq c$ is to be applied, even fractionally, then it must be the case that $v \uparrow \geq c$, otherwise the precondition on the action could never be met with any ordering of the actions chosen. Of course, if $v \uparrow \geq c$, we cannot guarantee that there *is* a legal ordering of the producers and consumers that achieves this value. This need to have at least a feasible opportunity to satisfy the precondition can be added to the LP through the introduction of a binary ($[0, 1]$ integer) variable and a pair of constraints. For the actions $[a_0..a_{n-1}]$ requiring a precondition $v \geq c$, then using $N$ to denote a large number, and $s$ to denote a new binary variable, we add the pair of constraints:

$$
\begin{aligned}
N.s &\geq a_0 + ... + a_{n-1} \\
v \uparrow &\geq lb(v) + (c - lb(v)).s
\end{aligned}
\tag{6}
$$

The first constraint forces $s$ to take the value 1 if any of the actions requiring the precondition $v \geq c$ are applied. The second constraint then determines the lower bound on $v \uparrow$ based on the value of $s$: if $s = 0$, then the lower bound on $v \uparrow$ is $lb(v)$, the global lower bound on $v$. Otherwise, $s = 1$, since an action needing the precondition has been applied, and thus $v \uparrow \geq c$. If there is no constraint that s = 1 then this implies that at least one of the actions $[a_0...a_{n-1}]$ is applied. However, since a non-zero value of $s$ only makes the LP harder to solve, there is no pressure to set $s = 1$ for any reason other than that the precondition must be satisfied. It is important to note that these constraints are only being added to the LP in problems where $v \geq c$ preconditions are not matched by a $v \mathrel{-}= c$ effect. The modified heuristic is able to support planning models that it previously could not. Its behaviour on domains without these characteristics is entirely unaffected.

## 8. Results

In this section we present a thorough evaluation of the LP-RPG heuristic: comparing it to state-of-the-art numeric planners, considering the use of different LP solvers and performing ablation studies to determine the most effective of the many potential different configurations of LP-RPG discussed in this paper. These include the weighting of action variables in the LP according to the RPG layer in which they appear, the inclusion of propositions and numeric goals in the LP and the consideration of which variables in the LP should remain integer and which should be relaxed to real numbers. All our tests were run on 3.4GHz Pentium IV machines and were limited to 30 minutes and 1.5GB of memory. If a planner or planner-configuration fails to report a solution within these limits it is deemed to have failed to solve the problem.

### 8.1 Evaluation Domains

First we discuss our selection of evaluation domains. Out purpose in selecting these domains is to select or construct examples that are informative in evaluating the behaviour of our heuristic. Domains with numeric variables that do not conform to the producer-consumer behaviour can be identified syntactically and a different planning strategy can then be employed. Since the syntactic analysis is trivial, the overhead in making this decision is negligible, so we can assume that the performance on domains to which our approach is not applicable is consistent with whichever alternative strategy is selected for deployment.

We consider both existing competition benchmarks with producer–consumer behaviour, and introduce some new domains that exhibit this behaviour. There are few current benchmarks that exhibit interesting producer–consumer behaviour, but in order to make a comparison that is as informative as possible we make use of those that do:

- the MPrime domain from IPC 1;

- the Rovers domain ('Numeric' variant) from IPC 3;

- the Settlers domain from IPC 3;

- an alternative encoding of the Settlers domain (described below);

- the Pathways domain from IPC 5. (We developed a metric domain derived from the 'Metric Time' variant, by replacing the durative actions with comparable non-temporal actions.)

In addition to the standard IPC problem set for Settlers (problems 1–20), we introduce some new problems that make full use of the scope of the domain. The domain allows for building ships and transporting materials between disjoint islands; however, in the benchmark set none of the problems require this. As building ships requires a large amount of infrastructure, we therefore add some problems to further challenge the planners, where materials must be imported from overseas in order to achieve goals. The first of our problems (21) requires merely the building of a ship, 22 requires the import of timber from overseas, and 23 requires building housing overseas. 24 adds some further goals to 23, requiring the planner to achieve both goals on the mainland and build housing on the island. The final problem, 25, considers 3 disjoint islands from which resources must be combined to achieve a goal on each island. Each of these problems requires building much greater infrastructure than required in the original IPC 3 settlers problems. We consider two variants of the settlers domain: the standard IPC 3 domain, and an encoding based on the representation of carts proposed by Gregory and Rendl (2008). Here, the number of carts at a given location is represented as a numeric variable (`carts-at ?location`). There are then two possible 'move' actions for carts: one that moves a cart and a unit of a resource from one location to another; and one that moves the cart without moving any resources, i.e. the cart moves whilst being empty. This encoding is possible because carts can only transport a single unit of material, so it is not necessary to maintain specific named identities and capacities for each cart.

In addition to the benchmark domains, we use two domains created during the development of lp-rpg:

- the Market Trader Domain (Coles et al., 2008);

- the Hydro Power Domain.

In the Market Trader Domain, a trader begins with a small amount of money, and the goal is to have increased that to a certain level. This must be achieved by travelling between markets, each of which sells some collection of goods at a certain price for each good type, and buys at another (lower) price. Money can be made by buying items where they are cheaper and transporting them (via camel) to other locations where they command a higher price; moving has an associated cost as food is required for the camel. This is a representation of a more general class of real-world trading problems where the aim is to make money through buying, transporting and selling goods. The Hydro Power domain is also concerned with financial gains, but the domain has a different structure: rather than transportation, it is concerned with energy storage using hydroelectric reservoirs. By buying electricity to pump water uphill during periods of low demand, when electricity is cheaper, and storing this potential energy, the electricity can be sold at a higher price at times of greater demand. The domain encoding is augmented to take into account energy loss: purchasing one unit of energy will not be sufficient to provide one unit of energy later as there are losses in the storage process. In this problem, despite the temporal axis, we are only interested in profit being made, so we do not force the planner to advance time to a specific point: we only ask that sufficient profit has been made and the planner can choose to advance time if necessary. In general this problem models continuous processes: customer demand changes continuously. Here, we simplify the problem by discretising into 30 minute time intervals, using the demand schedule from the transformer domain (Bell, Coles, Coles, Fox, & Long, 2009), based on the UK National Grid figures. Like the original transformer model we represent temporal features of the problem with an advance time action, rather than temporal PDDL, since LP-RPG is not a temporal planner.

| Domain | LP-RPG | LP-RPG-FF | MetricFF | LPG–td |
|---|---|---|---|---|
| Market Trader | 20 | 0 | 0 | 0 |
| Hydro Power | 27 | 3 | 1 | 5 |
| Pathways Metric | 30 | 21 | 13 | 0 |
| Settlers | 18 | 7 | 8 | 19 |
| Settlers Numeric Carts | 23 | 13 | 8 | 7 |
| MPrime | 30 | 30 | 28 | 30 |
| Rovers Numeric | 15 | 13 | 10 | 20 |
| Sugar | 18 | 7 | 14 | 20 |
| Total | 181 | 94 | 82 | 101 |

Table 2: Coverage achieved by different planners

The final domain we consider is the Sugar domain (Radzi, 2011). Here, the objective is to produce sugar through industrial processes, refining it from raw cane. This domain is taken from a set of domains designed for optimisation planning: in each there are several paths to the goal, originally included to allow the planner a choice of trajectories with the challenge being to find better quality solutions. As this set of domains was designed to be challenging as metric optimisation problems, most of the domains in the set are trivial for standard MetricFF: if the plan metric is ignored then almost all of the problems are solved in less than 1 second[6]. Therefore, of these domains, we consider only the Sugar domain, which remains challenging to MetricFF even when optimisation is not required: the number of paths that appear to lead to the goal is so large that without good guidance it is difficult to find any solution to the problem.

---

6. This is not to say that the other domains are uninteresting — they present an interesting challenge that is explored by Radzi (2011), using a carefully modified variant of LP-RPG. However, the challenge is to find good quality plans, where quality is determined by a more complex metric than simple plan length; neither MetricFF nor LP-RPG in the form discussed here have difficulty finding poor quality plans for these problems.

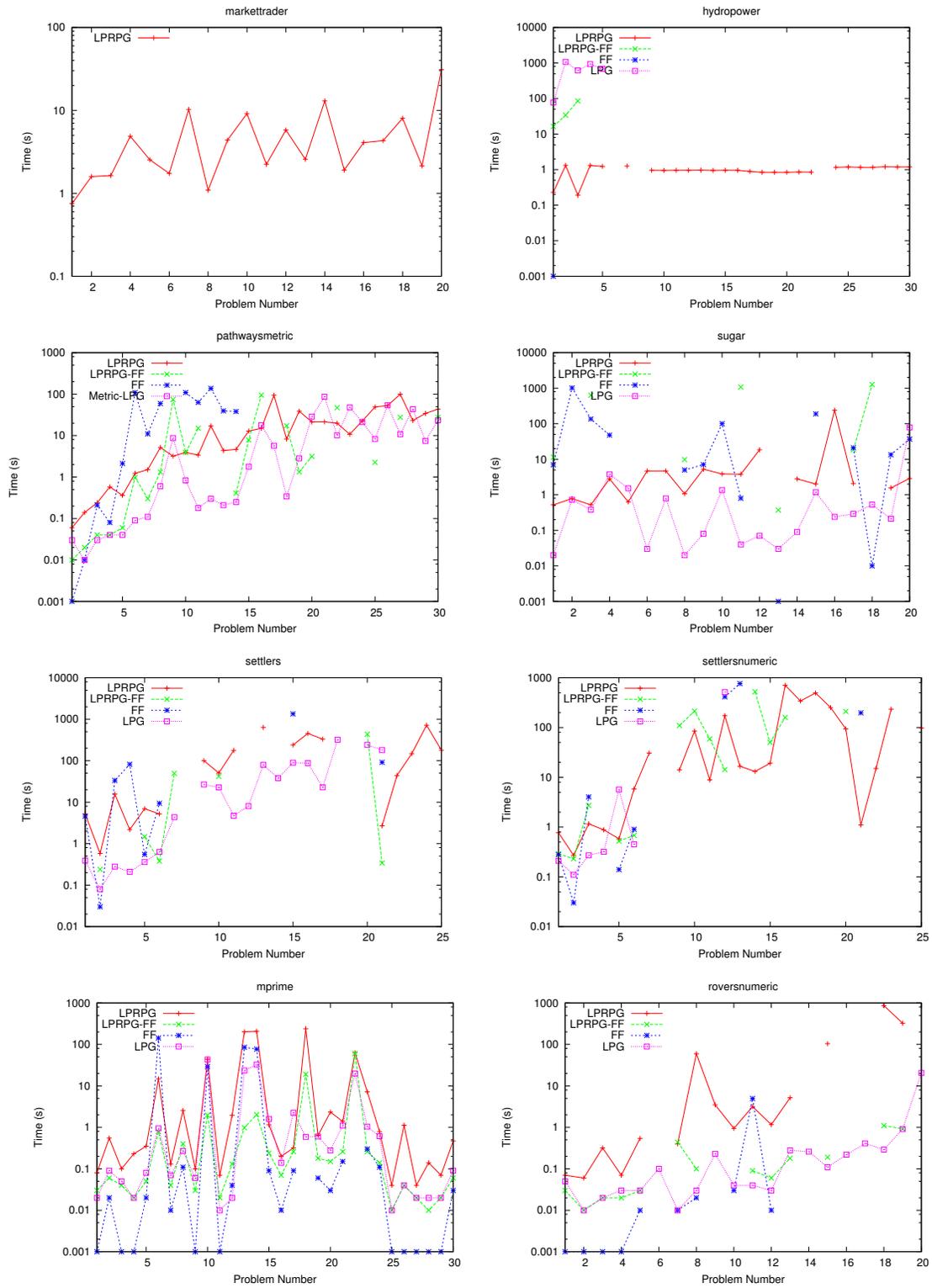

Figure 3: Comparison to MetricFF and LPG–td: time taken to solve problems

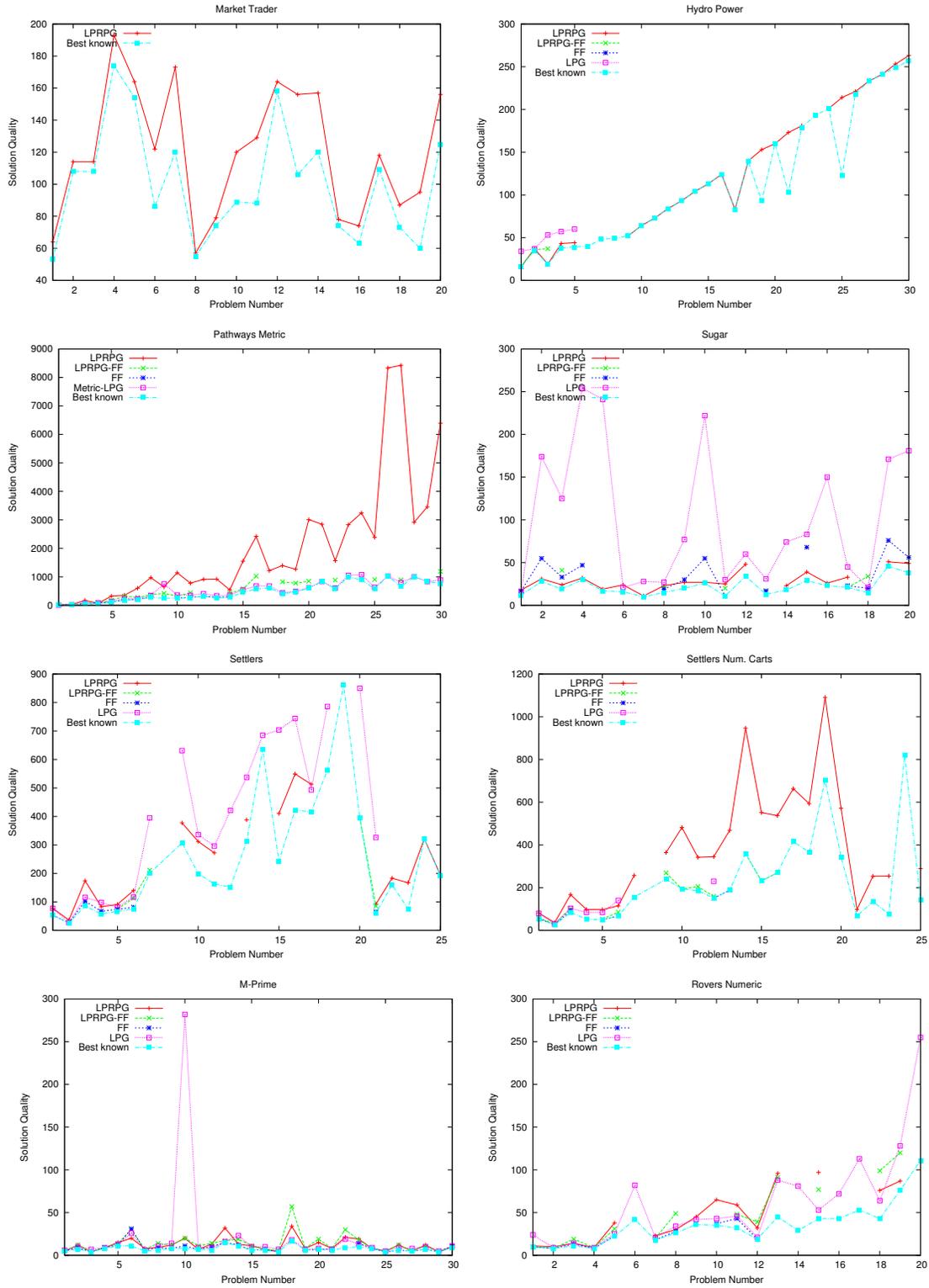

Figure 4: Comparison to MetricFF and LPG–td: plan length

## 8.2 Comparison to Other Planners

We first compare the performance of lp-rpg to existing numeric planners. We use what was found to be a strong (though not uniformly best) configuration of the planner, as we will demonstrate in subsequent sections:

- Landmarks and Propositional Goals are added to the LP (as in Section 6.1);

- The weight of an action variable in the objective function used during solution extraction is $3^l$, where $l$ is the layer at which it appeared during RPG expansion;

- Action variables corresponding to actions in action layer 1 are integral;

- IBM ILOG CPLEX version 12.1.0 is used as the LP solver.

We compare to the two historically most successful numeric planners: MetricFF (Hoffmann, 2003) and lpg–td (Gerevini et al., 2006). These remain state of the art, as many modern planners (e.g. lama) do not handle numeric preconditions, only action costs. To further clarify any differences in performance, we also compare to lp-rpg-FF: a reimplementation of MetricFF based on our lp-rpg code with the difference that, when computing upper- and lower-bounds on numeric variables during RPG expansion, lp-rpg-FF allows actions to be applied many times at the same action layer, rather than only once per action layer as in MetricFF. Since the publication of our earlier comparison to lpg–td (Coles et al., 2008), a new and improved version of lpg–td has been produced. This version of lpg–td performs much better than the earlier version, and we use it in our results here. We also consider a variant of lpg–td, Metric-lpg–td (Gerevini, Saetti, & Serina, 2008), designed to be more responsive to plan metrics based on numeric variables. Our experiments showed that Metric-lpg–td does not perform significantly differently to lpg–td in generating first feasible plans for problems where plan length is the metric, apart from in the Pathways domain where lpg–td crashes on all problems. Therefore, we report performance for lpg–td in all domains except Pathways where we report figures for Metric-lpg–td.

An interesting pattern emerges in the relative performance of the planners across our set of evaluation domains, shown in Figure 3. The domains are organised with those at the top being the most strongly numeric, relying on few propositions, and those towards the bottom having more propositional structure and consequently less numeric structure. On the two most heavily propositional domains (MPrime and Rovers) lpg–td is generally the most successful planner, solving all of the problems in the evaluation sets, and often being the fastest planner on these problems. The same pattern holds for the standard competition problems (1-20) in the competition formulation of the Settlers domain. MetricFF also performs quite well on the MPrime and Rovers domains, but struggles on the Settlers domain due to the numeric structure present.

In the more strongly numeric domains, however, lpg–td performs poorly: indeed it fails to solve a single problem in the Pathways and Market Trader domains. This is not due to crashing, rather the planner just searches until it exhausts resource limits without finding a solution. In Hydro Power, lpg–td solves the five easiest problems, but is not able to scale beyond this. In our experiments we have observed that lpg–td struggles on domains where there is limited propositional structure more generally, and the search guidance it gets from the numeric problems is poor. Comparing the two Settlers variants also gives some interesting insights: when the carts are turned in to a numeric resource, LPG–td struggles much more solving 7, instead of 19 problems, whereas the performance of LP-RPG is in fact improved. Note that although LPG–td is successful on the IPC 3 problems, it cannot solve the richer problems where ship building and overseas transport is required; whereas LP-RPG is capable of solving such instances.

Turning our attention to the comparison with MetricFF we can observe that problems solved by both planners are generally solved more quickly by MetricFF, particularly on the domains with more propositional structure. This is because LP-RPG has the additional overhead of solving an LP at each state (and partly, of course, due to the highly efficient MetricFF code-base). Occasionally the general pattern is broken, with LP-RPG being faster; this is because slight variations in the ordering of heuristically equivalent states can lead to significant differences in performance. The results for LP-RPG-FF show similar over all coverage to those for MetricFF, although sometimes solving different problems (again branch orderings and a different code base can cause such differences, which are most marked in the Pathways and Sugar domains), but serve to demonstrate that it is not our basic FF implementation performing drastically differently to standard MetricFF that is causing the gains we observe.

Looking at the numeric domains in particular, the LP-RPG heuristic is able to provide much better guidance, allowing LP-RPG to solve many more problems than MetricFF. Notably, in the Market Trader domain neither MetricFF nor LP-RPG-FF can solve any problems. This is due to the poor guidance the standard RPG heuristic gives in this domain, relaxing the transfer of numeric resources. The relaxed plan is to buy an item then repeatedly sell the same item until sufficient profit has been made. Again, in Hydro Power, a similar situation occurs: once one unit of energy has been pumped up, the same unit of energy can be repeatedly sold at any future time of day, making sufficient profit without any guidance. In Pathways, where chemical reactions must take place, the relaxation used by MetricFF will allow units of the same substance to be used repeatedly, in several different reactions, despite the fact that when they are used they are consumed. The LP-RPG heuristic does not permit this and therefore gives much better search guidance, allowing LP-RPG to solve all problems in this domain. The numeric resource transfer present in the Settlers domain leads to poor guidance from the MetricFF heuristic and MetricFF is able to solve very few problems as a result. The use of the LP is very effective in this domain allowing more problems to be solved. The different formulations seem to make little difference to coverage for MetricFF, with neither making the problems easier for MetricFF to solve.

The quality of solutions (plan length) produced by the different planners is displayed in Figure 4. We emphasize here that LP-RPG in its current form is not making any attempt to minimise a general measure of plan quality, these results are merely intended to give an indication as to whether there is a large degradation, or indeed a fortuitous increase, in quality in moving from using a standard RPG heuristic to the hybrid LP-RPG approach. The potential to improve plan quality using an LP-RPG-style approach has been explored in other work for domains with preferences (Coles & Coles, 2011) and also for a range of different metrics (Radzi, 2011). None of the problems we use have specified metric functions to minimise, so instead we use the number of actions in the solution plan. This is the value that the RPG heuristic will tend to minimise. None of the planners are run in optimisation mode (where that is available) and all simply report the first plan found in search. On most domains the quality of solutions produced by LP-RPG is comparable to that of those produced by MetricFF and LPG–td. In the sugar domain the LP-RPG heuristic compares favourably to LPG–td, although we could perhaps hope that running LPG–td in quality mode would enable it to produce better solutions. In Pathways, LP-RPG produces particularly long solutions, but there is a trade off, as it is also able to scale to solve far more problems. We will return to the issue of solution length in this domain when considering the weighting of action variables in the LP during solution extraction.

In summary, LPG–td seems to be generally successful on domains where there is sufficient propositional structure and MetricFF is generally very efficient on problems that it is capable of solving. When the structure of the domain becomes heavily numeric both of these planners perform poorly. LP-RPG, however, is able to solve many more problems than any of the other planners, making use of search guidance in the LP that captures numeric interactions well.

### 8.3 LP Solvers

In LP-RPG the construction and use of an LP is performed using functions commonly available in a wide range of LP solvers: adding variables or constraints to a model, setting variable bounds, marking whether variables are real or integer valued, changing the objective function, and so on. The current implementation employs a minimal abstraction layer between LP-RPG and the LP solver itself, so that almost any LP solver can be used. In this section, we consider the use of three LP solvers:

- IBM ILOG CPLEX version 12.1.0, a commercial mixed integer-linear programming solver.

- COIN-OR LP (CLP) version 1.12.0, an open-source LP solver. Where models feature integer variables, CLP is used within COIN-OR Branch-and-Cut (CBC) version 2.5.0, which is, again, open-source.

- LPSolve version 5.5.0.13, an open-source mixed integer-linear programming solver.

In our experiments with these LP solvers, we found that CPLEX is substantially more robust than the other two, particularly when the LP is extended to include satisfying propositional goals and landmarks. Thus, for the purposes of the comparison here, we will use a configuration of LP-RPG (equivalent to that used in our earlier paper (Coles et al., 2008)) that is not as efficient as those presented elsewhere in the paper, but is most robust (caused CLP and LPSolve to crash less often) under testing:

- Propositional goals and landmarks are *not* added to the LP: it encodes numeric goals only.

- The only integer variables are (potentially) helpful actions, or those with assignment effects.

- The layer-weighting scheme with $k = 1.1$ is used.

We refer to this configuration as limited-LP-RPG.

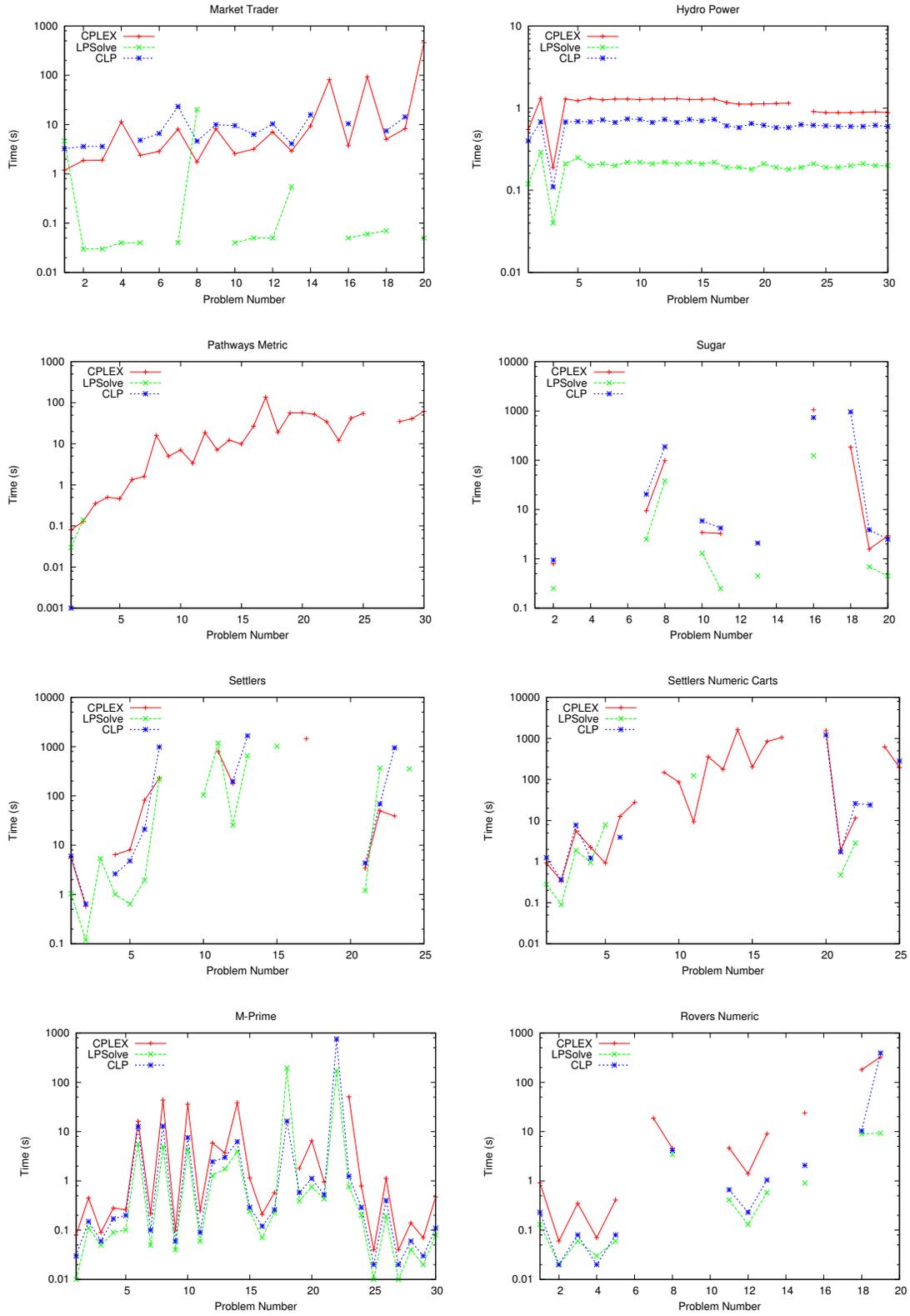

Figure 5: Time taken by limited-LP-RPG to solve problems using different LP solvers

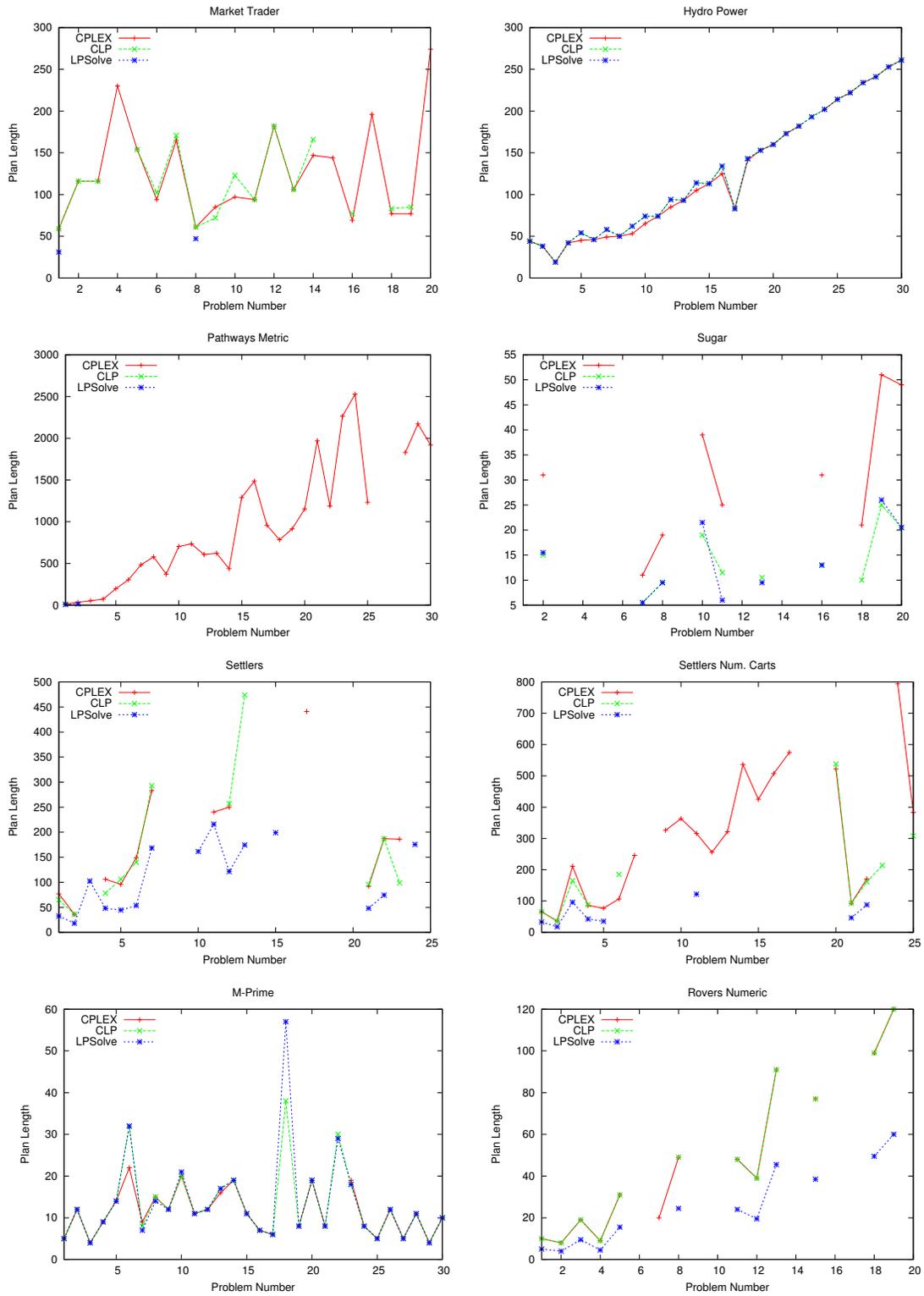

Figure 6: Lengths of plans produced by limited-LP-RPG using different LP solvers

| Domain | CPLEX | CLP | LPSolve |
|--------|-------|-----|---------|
| Market Trader | 20 | 16 | 15 |
| Hydro Power | 29 | 30 | 30 |
| Pathways Metric | 28 | 1 | 2 |
| Sugar | 9 | 10 | 9 |
| Settlers | 12 | 11 | 15 |
| Settlers Numeric Carts | 21 | 10 | 8 |
| MPrime | 28 | 30 | 30 |
| Rovers Numeric | 13 | 12 | 12 |
| Total | 160 | 120 | 121 |

Table 3: Coverage of limited-LP-RPG with different LP solvers

The results of these tests are shown in Figure 5, and Table 3. Beginning with the Market Trader domain, it is quite clear that CPLEX is faster than CLP in this domain. LPSolve, in turn, substantially out-performs CPLEX on many problems, by two orders of magnitude. This speed, though, comes at a cost in terms of robustness: CPLEX solves all 20 problems, but LPSolve solves only 15. LPSolve also demonstrates strong performance in the Hydro Power domain, while CLP falls between LPSolve and CPLEX.

The Pathways Metric domain illustrates the robustness of CPLEX to extension beyond a basic producer–consumer model encoding. This domain contains actions with numeric preconditions that must be true, but are not affected by the action. As described in Section 7.2, encoding this requires an integer variable for each such precondition. This domain also contains goals referring to multiple numeric variables, e.g:

```
(>= (+ (available cycDp1) (available c-Myc-Max)) 3)))
```

As can be seen, in this domain, CPLEX is the only LP solver which allows LP-RPG to solve anything but the smallest problems. Beyond problem 2, CLP is unable to solve the LP to reach the goals from the initial state. Using LPSolve, solvable LPs are reported unsolvable, so whilst the planner can make an attempt at search, erroneous state pruning happening as a result of LPs being falsely declared unsolvable renders it unable to find solutions.

In the Sugar domain, no solver leads to the planner performing particularly well, with no more than 10 problems being solved. This contrasts with earlier results, shown in Table 2, where a different configuration of LP-RPG, using CPLEX and richer LPs, was able to solve 18 problems. However, as noted at the start of this section, we have had to compromise the performance of the planner when using CPLEX to allow a reasonable comparison with CLP and LPSolve: using the richer LP models here, CPLEX solves 18 problems but LPSolve and CLP both perform far worse (falsely claiming LPs are unsolvable, or returning suboptimal solutions, to the detriment of the performance of the planner).

In the two different encodings of the Settlers domain, where carts are represented either explicitly or using the `carts-at` function, we can see how CPLEX is more robust under alternative domain encodings. In the original IPC domain model, LPSolve performs particularly well, while CLP is not markedly different to CPLEX. Using the numeric-carts encoding, however, CPLEX is considerably better than the other solvers, being robust to the change in the LP structure arising from this alternative domain encoding.

| Domain | LPSolve | | CLP | | CPLEX | |
|---|---|---|---|---|---|---|
| | Build (ms) | Solve (ms) | Build (ms) | Solve (ms) | Build (ms) | Solve (ms) |
| Market Trader | 5.0 | 167.1 | 1.3 | 137.9 | 17.3 | 44.6 |
| Hydro Power | 0.7 | 1.4 | 0.3 | 5.9 | 9.4 | 6.6 |
| Pathways Metric | 0.0 | 1.3 | 4.3 | 8.8 | 4.4 | 3.1 |
| Sugar | 0.8 | 7.5 | 1.6 | 42.0 | 17.6 | 22.9 |
| Settlers | 2.6 | 31.9 | 2.2 | 170.6 | 165.5 | 87.6 |
| Settlers Num. Carts | 2.2 | 22.9 | 1.6 | 71.0 | 48.0 | 33.6 |
| MPrime | 3.1 | 5.4 | 5.3 | 11.5 | 76.2 | 6.3 |
| Rovers Numeric | 0.8 | 4.6 | 0.7 | 7.2 | 14.8 | 1.9 |
| Average | 1.9 | 30.3 | 2.2 | 56.8 | 44.1 | 25.8 |

Table 4: Time spent in LP building and solving using different LP solvers

In the MPrime and Rovers domain, CLP and LPSolve are consistently faster than CPLEX.

Summarising, we can see that if limited-LP-RPG can solve a problem using CLP or LPSolve, it is usually faster than using CPLEX, but CPLEX offers better coverage and its greater robustness grants access to the richer encodings that allow better performance.

In order to investigate in more detail why the planner takes much longer to solve problems using CPLEX we devised some further tests. Table 4 shows the average time spent, per state, building and solving the LP for each LP solver. (The building time is the time required to integrate constraints inserted by LP-RPG into the internal model used by the particular solver.) Note that these results are not necessarily directly comparable: the planners do not necessarily take the same paths through the state space, so might be evaluating different states. The paths are, nonetheless, often similar and the times can be taken as strongly indicative. To give the fairest possible comparison, we include in this table only data for problems that were solved by all three configurations, so the data is presented for each planner across exactly the same problem set. A startling observation is that CPLEX typically spends an order of magnitude longer building LPs than the other two solvers (in Settlers two orders of magnitude). So, while it often solves the LPs more quickly, the total time spent handling the LPs is generally greater. Indeed, for CPLEX, the time spent solving the LPs is, in 5 out of 8 domains, dominated by time spent building them. This indicates that, although CPLEX should be a good choice to use as the LP solver (it solves the LPs efficiently), in practice other solvers are faster due to the substantial overheads of building the large number of LPs necessary (at least one per state). These results suggest that a robust LP solver with low LP building overheads could dramatically improve the performance of LP-RPG.

When considering solution quality we note that the only way in which the LP Solvers can direct search on to a different trajectory is if the solvers return different optimal solutions to the same LP at some point during search. Recall that no planner configuration seeks to directly minimise plan length: the objective function in the LPs uses a weighted sum of the number of actions. Figure 6 shows that a similar picture arises in plan length as in time performance: LPSolve often leads the planner to shorter solutions than CPLEX (also helping to explain why it is often faster, since it explores the search tree to a smaller depth). In particular this happens in the Rovers Numeric and Sugar domains as well as the Settlers variants. Across other domains there is little variation in the quality of solutions produced.

| Domain | $1^l$ | $1.1^l$ | $3^l$ | $5^l$ | $10^l$ | $h_{max}$ | $h_{add}$ |
|---|---|---|---|---|---|---|---|
| Market Trader | 18 | 20 | 20 | 20 | 20 | 20 | 20 |
| Hydro Power | 30 | 29 | 27 | 27 | 26 | 28 | 28 |
| Pathways Metric | 30 | 28 | 30 | 30 | 30 | 30 | 30 |
| Sugar | 11 | 9 | 18 | 18 | 20 | 16 | 16 |
| Settlers | 13 | 19 | 18 | 17 | 16 | 20 | 22 |
| Settlers Numeric Carts | 22 | 22 | 23 | 21 | 21 | 23 | 23 |
| MPrime | 29 | 30 | 30 | 30 | 30 | 30 | 29 |
| Rovers Numeric | 10 | 13 | 15 | 15 | 15 | 14 | 15 |
| Total | 163 | 170 | 181 | 178 | 178 | 181 | 183 |

Table 5: Coverage when varying LP objective function weighting schemes

## 8.4 LP Objective Function Weighting Schemes

When using the LP during the solution extraction phase, one open issue is what weighting scheme to use in the objective function. Since the LP ignores the propositional preconditions of actions, using the simple objective of minimising the sum of action variables (layer-weighting scheme with $k = 1$) gives the LP solver freedom to select equally between actions appearing in any layers of the RPG, regardless of how many actions subsequently need to be added to the relaxed plan to support them. As discussed earlier, using a layer-weighting scheme with $k > 1$, or using a weighting scheme based on estimated costs of achieving the preconditions of the actions, should encourage the LP solutions to favour actions that are cheaper to apply. The $h^{max}$ or $h^{add}$ heuristics are both candidates on which to base estimates for the cost of application of actions. Of course, there will be cases in which the choice of an earlier action, or of one that has lower costs to achieve its preconditions, will be a flawed choice, worse than the choice that would have been made using the $k = 1$ layer-weighting scheme: that is simply the nature of heuristics.

In this section, we evaluate a range of LP action-variable layer-weighting schemes and also action-cost estimate schemes. We use $k \in 1, 1.1, 3, 5, 10$ for the layer-weighting schemes. We consider setting action variable weights to 1 plus the cost of meeting their propositional preconditions with $h^{max}$ or $h^{add}$ for the action-cost estimate schemes. The other parameters of the planners are set to sensible defaults: action variables in the first action layer are integral, and propositional goals and landmarks are included in the goal-checking LP.

Results showing coverage of each of these configurations are shown in Table 5. Our first observation is that *any* other weighting scheme is better than using a layer-weighting with $k = 1$ for action variables. This is particularly noticeable in the Settlers domain, where there are several situations in which the earlier actions should be preferred. For example,

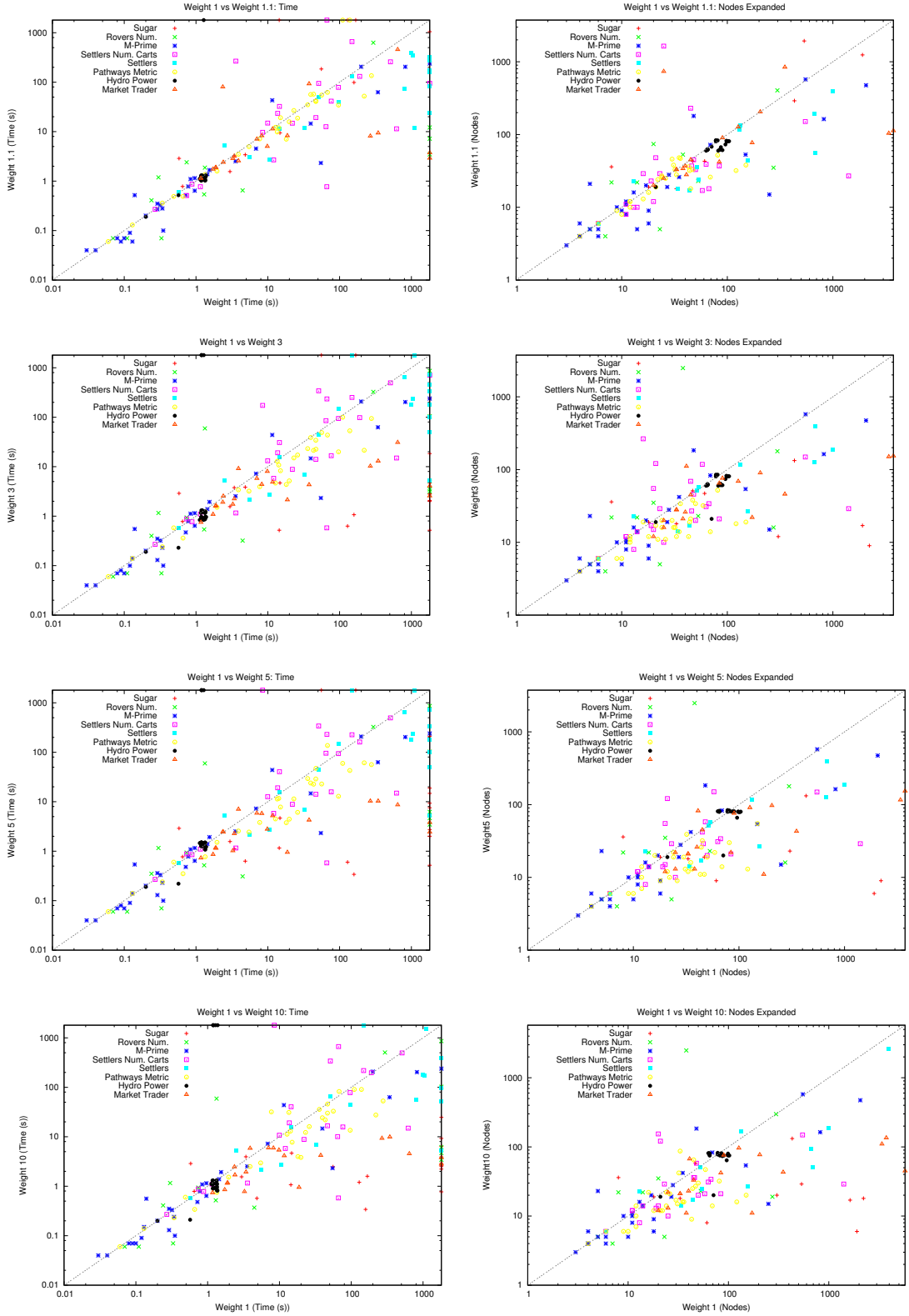

Figure 7: Layer-weighting schemes in the LP ($k = 1$ versus $k = x$ for different values of $x$)

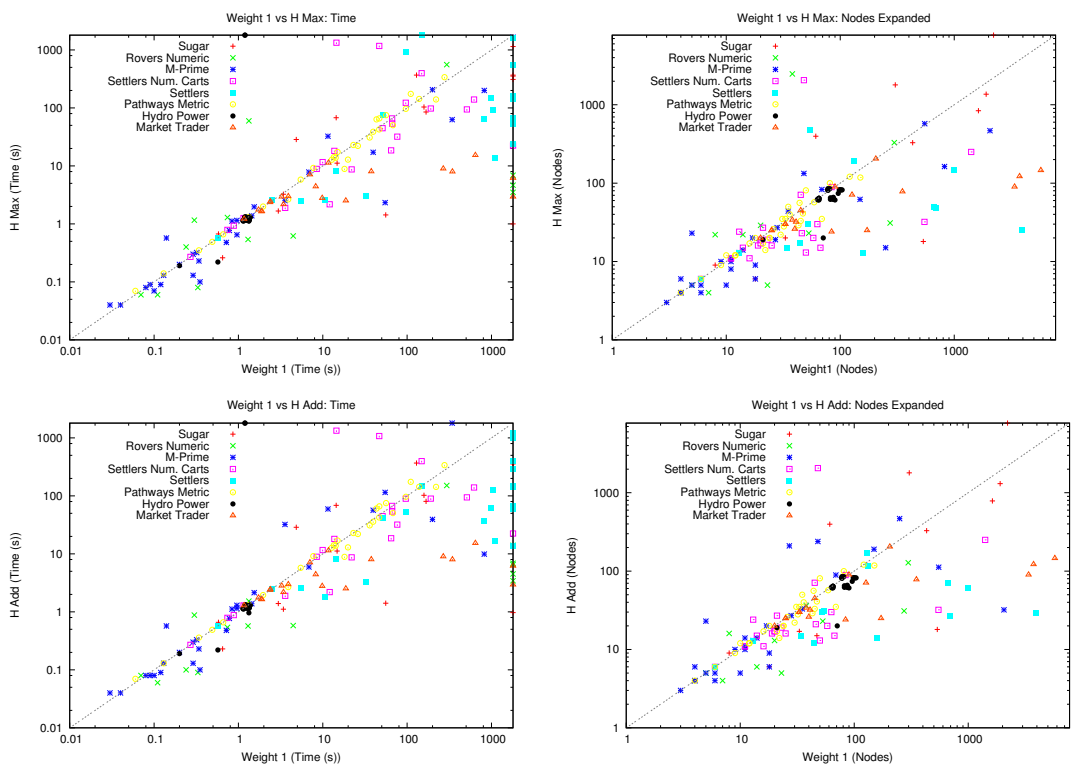

Figure 8: Action weighting schemes in LP objective function (layer-weighting with $k = 1$ versus action-cost estimate based schemes using $h_{add}$ or $h_{max}$)

supposing a cart is at location $A$, and a unit of a resource, initially available at both $A$ and $B$, needs to be moved to location $C$. There are two two-action solutions to the LP. Annotating each action with the layer at which it appears, these are:

- load cart at $A$ (0), unload cart at $C$ (1);

- load cart at $B$ (1), unload cart at $C$ (1).

Using the layer-weighting scheme with $k = 1$ these two solutions are indistinguishable (each has cost 2). However, if the latter is selected, two actions are needed as support in the relaxed plan: moving the cart to $B$ and moving the cart to $C$. Any of the alternative action variable weighting schemes set the weight of loading the cart at $B$ to be higher than that of loading it at $A$, leading to a preference for the first solution, which is a better outcome for search guidance. A similar phenomenon occurs in the Rovers domain, where favouring earlier actions increases the preference for recharging the rovers at or close to their current locations. In this domain this leads to better avoidance of dead-ends, since postponing recharging actions risks the possibility that the rover will have too little charge left to get back to a recharging location.

The data for the layer-weighting schemes show that the results peak at $k = 3$, a reasonable trade-off between minimising the number of actions chosen in the LP solution and favouring earlier actions. The overall performance of $h^{max}$ is the same, losing performance in Sugar and Rovers but gaining in Settlers and Hydro Power. Using $h^{add}$ gives further gains in Settlers, leading to very slightly better coverage than the layer-weighting scheme with $k = 3$. The difference in performance between $k = 3$ and $h^{add}$ weighting schemes is domain-dependent. $h^{add}$ only increases the weight of an action above 1 if it has propositional preconditions which are not true in the state being evaluated. In the Settlers domain, where it gives particularly good performance, this is a sound approach, neatly capturing the example case discussed earlier in this section: loading or unloading from a cart at its current location is preferable to doing so at later locations. In domains where earlier actions are preferable even if their propositional preconditions require little or no support, $h^{add}$ fails to give an adequate bias, and the layer-weighting scheme with higher $k$ performs better. For instance, in the Sugar domain, the best coverage is obtained by using $k = 10$, and the $k = 3$ scheme also performs more strongly than $h^{add}$ here.

| | Time Taken | | | | | | Nodes Expanded | | | | | | Plan Length | | | | | |
|---|---|---|---|---|---|---|---|---|---|---|---|---|---|---|---|---|---|---|
| $k =$ | 1.1 | 3 | 5 | 10 | $h_{max}$ | $h_{add}$ | 1.1 | 3 | 5 | 10 | $h_{max}$ | $h_{add}$ | 1.1 | 3 | 5 | 10 | $h_{max}$ | $h_{add}$ |
| $k = 1$ | ✓ | ✓ | ✓ | ✓ | ✓ | ✓ | ✓ | ✓ | ✓ | ✓ | ✓ | ✓ | ✗ | ✓ | ✓ | ✓ | ✓ | ✓ |
| $k = 1.1$ | | ✓ | ✓ | ✓ | ✗ | ✗ | | ✓ | ✓ | ✓ | ✗ | ✓ | | ✓ | ✓ | ✓ | ✓ | ✓ |
| $k = 3$ | | | ✗ | ✗ | ✓ | ✓ | | | ✗ | ✗ | ✓ | ✓ | | | ✓ | ✓ | ✓ | ✓ |
| $k = 5$ | | | | ✓ | ✓ | ✓ | | | | ✗ | ✓ | ✗ | | | | ✗ | ✓ | ✓ |
| $k = 10$ | | | | | ✓ | ✓ | | | | | ✓ | ✗ | | | | | ✓ | ✓ |
| $h_{max}$ | | | | | | ✗ | | | | | | ✓ | | | | | | ✗ |

Table 6: Results of Two-Tailed Wilcoxon Signed Rank Tests comparing different LP weighting schemes. ✓ indicates significance ($0 = 0.05$), colour indicates the better performer (faster, fewer nodes expanded or shorter plans).

Examining the performance of the configurations in more detail, comparing the time taken to find solution plans and the number of nodes evaluated, scatterplots comparing each of the configurations tested to layer-weighting with $k = 1$ are shown in Figures 7 and 8, for other layer-weighting schemes and action-cost estimate based schemes. As LP-RPG usually exhausts the time limit of 30 minutes before the memory limit, the time-taken scatterplots are closely similar to the coverage table. Coverage is directly reflected in the number of points on the far right of the graphs, indicating where a layer-weighting scheme with $k = 1$ (the x-axis) was unable to find a solution within 30 minutes, but LP-RPG using a different weighting scheme was able to solve the problem. There are more such points for $k = 3$ and $h^{add}$ than the other schemes, with the points for $h^{add}$ appearing predominantly in the Settlers domain and those for $k = 3$ spread across the domains.

Since it is not always clear from the scatterplots whether one configuration is better than the other or whether the differences are significant, we have used the Two-Tailed Wilcoxon Signed Rank Test to compare each pair of tested configurations in terms of time taken and nodes evaluated, and also plan length.[7] All tests are performed with $p = 0.05$. The results of these tests are summarised in Table 6. A number of interesting observations can be made:

---

7. The Wilcoxon signed-rank test is a non-parametric statistical test used to compare a set of matched samples (such as the pairs of results for two different planners on the same sequence of problems) to assess whether their population mean ranks differ (i.e. it is a paired difference test). It is useful when the absolute values are not necessarily comparable and when the samples are drawn from a completely

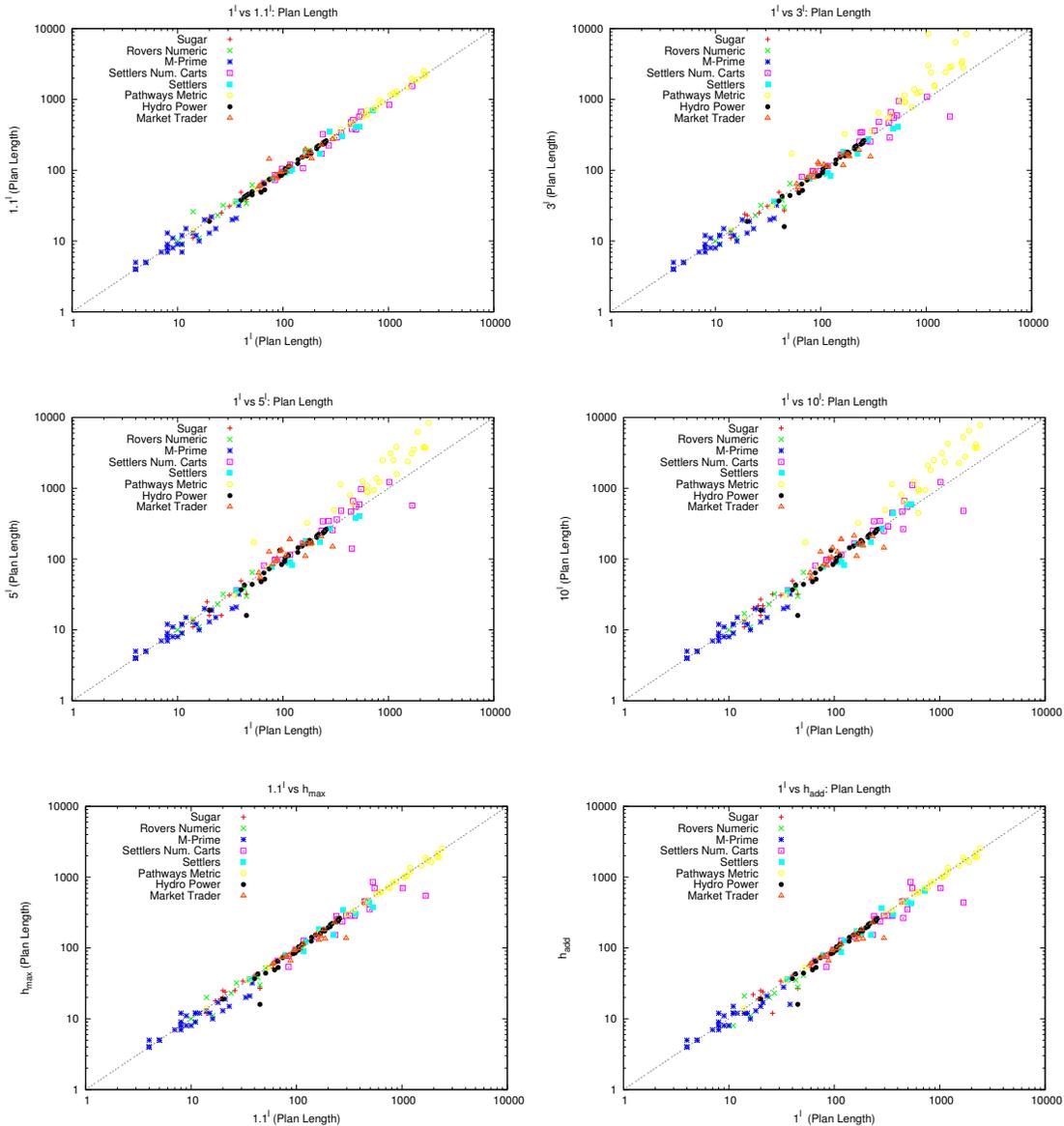

Figure 9: Effect on quality of varying action weighting schemes in LP objective function

- While $k = 3$ gives better coverage than $k = 5$ or $k = 10$, it does not take a significantly different amount of time nor explore a significantly different number of nodes.

- There is, however, a significant difference in time taken and nodes, in favour of $k = 3$, when compared to all the other configurations tested.

- The action-cost based weighting schemes are slower than the layer-weighting schemes for $k \geq 3$. Furthermore, when compared to $k = 3$, they expand more nodes.

unknown distribution. The two-tailed version makes no assumption about which of the two contenders has the better performance (i.e. the lower mean rank).

- The key strength of the action-cost based weighting schemes lies in the quality of the solutions found. As a general trend, layer-weighting schemes gain coverage but lose quality as $k$ increases. The action-cost based schemes generate significantly shorter plans than any other configuration tested, including layer-weighting with $k = 1$.

The increase in plan length for plans found using the layer-weighting scheme with increasing $k$ can be seen in Figure 9. The reason for this increase is quite clear: the basic objective function for the LP is to minimise number of actions. However, when weightings are added to the actions according to the RPG layer in which they appear, longer plans that make use of earlier actions can become more attractive. The increase in plan length is, however, largely restricted to two domains. The first, and worst affected, is Pathways Metric, where the increase in $k$ causes a preference for reactions that are less efficient (in terms of the number of actions needed) but can be performed using actions from the earliest layers of the planning graph. The second is Settlers with Numeric Carts, where the increase in $k$ leads again to solutions with a preference for resource production and refinement approaches that are less efficient, but comprise actions that appear earlier in the planning graphs. As is perhaps to be expected, when these two domains are excluded from the statistical analysis, there is no significant difference in plan length between $k = 3$ and the action-cost based schemes.

Of the weighting schemes tested the best results are obtained using $k = 3$ and $h^{add}$: there is no single best option. The former is faster, expands fewer nodes, and gives a good balance in coverage across the domains used. The latter is less prone to variations in plan quality in some domains, and its strong performance in the Settlers domain leads to two more problems being solved overall within the test domains.

## 8.5 The Use of Integer Constraints

In Section 5.4.1 we discussed the fact that the LP is a relaxation of a MIP, and proposed that in certain situations it may be beneficial to not relax some action variables, making them integral. In this section, we will explore this hypothesis, considering five configurations of LP-RPG:

1. Minimal Integers: only actions with assignment effects (as in Section 4.3) are integers.

2. First-layer: as above, but variables for actions appearing in RPG layer 1 (the potentially helpful actions) are also integral.

3. Propositional-Goal Achievers: as above, but the variables for any actions that achieve propositional goals or landmarks are also integral.

4. Numeric-Goal Achievers: as above, but additionally, the variables for actions affecting numeric state variables that appear in numeric goals are also integral.

5. All: all (action) variables are integral.

The coverage of each of these configurations is shown in Tables 7 and 8, for layer-weighting using $k = 1.1$ and $k = 3$, respectively. For the $k = 3$ results, the coverage is fairly insensitive to the configuration used. There is a peak in coverage with the 'Numeric Goal

| Domain | Minimal (Assignments) | First-Layer Actions | Prop. Goal Achievers | Num. Goal Achievers | All Variables |
|---|---|---|---|---|---|
| Market Trader | 20 | 20 | 20 | 20 | 20 |
| Hydro Power | 23 | 29 | 29 | 29 | 29 |
| Pathways Metric | 30 | 28 | 28 | 30 | 29 |
| Settlers | 21 | 19 | 19 | 19 | 19 |
| Settlers Num. Carts | 22 | 22 | 22 | 22 | 21 |
| MPrime | 30 | 30 | 30 | 30 | 30 |
| Rovers Numeric | 14 | 13 | 13 | 13 | 15 |
| Sugar | 10 | 9 | 9 | 9 | 19 |
| Total | 170 | 170 | 170 | 172 | 182 |

Table 7: Coverage varying which action variables are integer in the MIP (layer-weighting with $k = 1.1$)

| Domain | Minimal (Assignments) | First-Layer Actions | Prop. Goal Achievers | Num. Goal Achievers | All Variables |
|---|---|---|---|---|---|
| Market Trader | 20 | 20 | 20 | 20 | 20 |
| Hydro Power | 23 | 27 | 27 | 29 | 29 |
| Pathways Metric | 30 | 30 | 30 | 30 | 30 |
| Settlers | 20 | 18 | 18 | 18 | 15 |
| Settlers Num. Carts | 22 | 23 | 23 | 23 | 23 |
| MPrime | 30 | 30 | 30 | 30 | 30 |
| Rovers Numeric | 15 | 15 | 15 | 15 | 15 |
| Sugar | 20 | 18 | 18 | 18 | 20 |
| Total | 180 | 181 | 181 | 183 | 182 |

Table 8: Coverage varying which action variables are integer in the MIP (layer-weighting with $k = 3$)

Achievers' configuration, though the difference between that and the worst configuration is only 3 problems. Using $k = 1.1$, there is a marked increase in coverage when all action variables are integral. This is due to the Sugar domain: compared to the preceding configuration in the table, an additional 10 problems are solved. In the same domain, when using $k = 3$, though, even better coverage is obtained when using only the *minimal* number of integral action variables. Thus, it appears the need for integral variables in this domain is reduced once the objective preference for earlier actions is sufficiently high. Disregarding the Sugar domain, the results for $k = 1.1$ in other domains are very close, as in the $k = 3$ configuration.

It is interesting that, even with many integers in the MIP, the performance of LP-RPG in terms of coverage is not very different from using no integers at all. Considering the computational complexity of solving a MIP, rather than solving an LP, the time spent calculating the heuristic should be considerably higher, rendering an all-integers approach impractical. Investigating this further, scatterplots of the time taken and the number of nodes expanded are shown in Figure 10. (The results shown are for $k = 3$, but the overall picture is the same if $k = 1.1$ is used.) Each configuration is compared to the 'Minimal Integers' configuration. The general trend, as one moves down the left column (increasing the proportion of action variables that are integral), is for more points to drift above the line $y = x$, i.e. the time taken to solve problems increases. At the same time, though, moving down the right-hand column there is a decrease in the number of nodes evaluated. Thus, increasing the proportion of integral action variables seems to improve search guidance, though not sufficiently to allow a pay-off in terms of the time taken to solve problems.

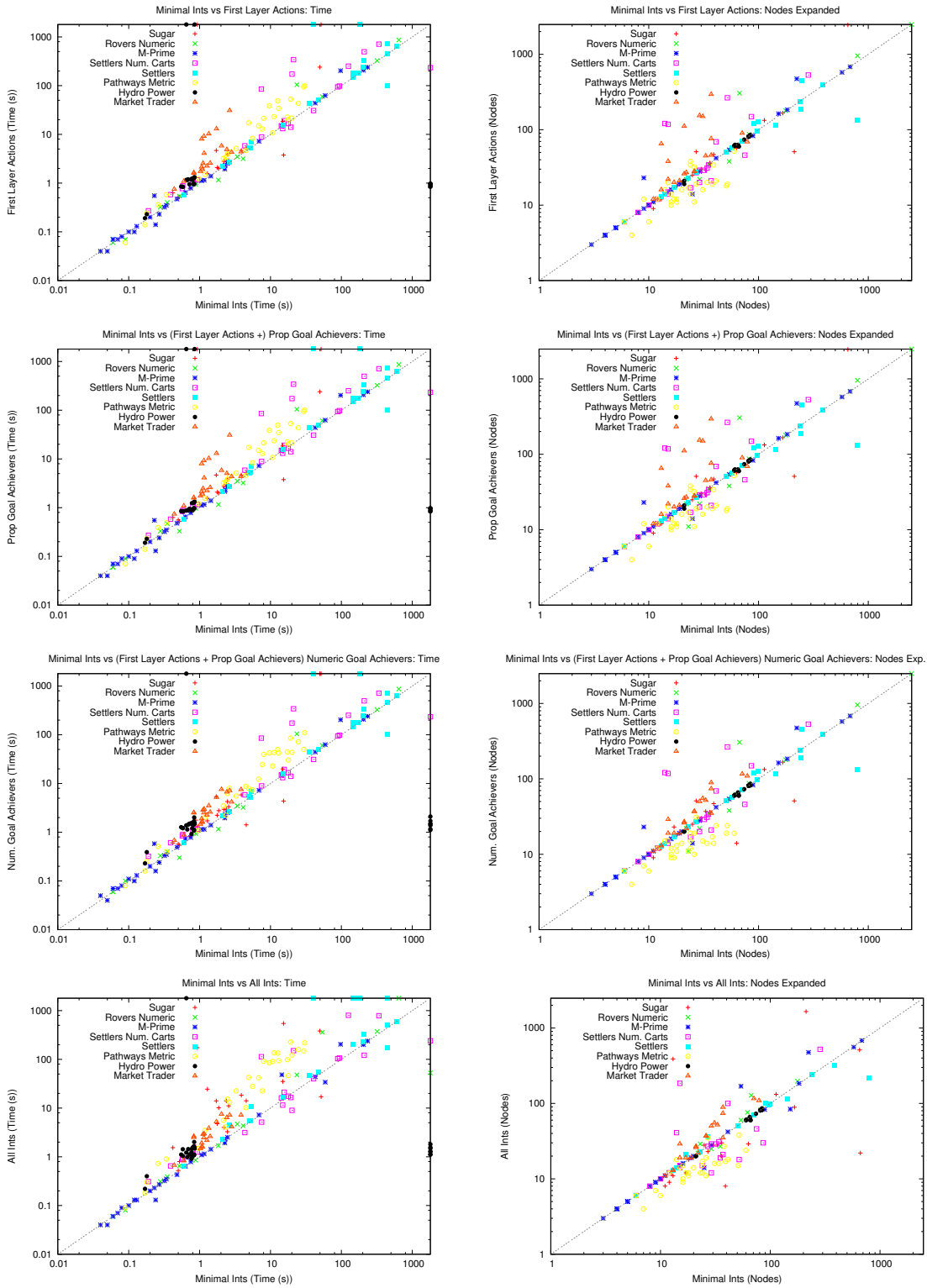

Figure 10: Changing which variables are integer ($k = 3$)

| | Time Taken | | | | Nodes Expanded | | | | Plan Length | | | |
|---|---|---|---|---|---|---|---|---|---|---|---|---|
| | First Layer Acts | Prop. Goal Ach. | Num. Goal Ach. | All Ints | First Layer Acts | Prop. Goal Ach. | Num. Goal Ach. | All Ints | First Layer Acts | Prop. Goal Ach. | Num. Goal Ach. | All Ints |
| Minimal Ints | ✓ | ✓ | ✓ | ✓ | ✗ | ✗ | ✗ | ✗ | ✗ | ✗ | ✗ | ✗ |
| First Layer Acts | | ✗ | ✓ | ✓ | | ✗ | ✓ | ✓ | | ✗ | ✗ | ✗ |
| Prop. Goal Ach. | | | ✓ | ✓ | | | ✓ | ✓ | | | ✗ | ✗ |
| Num. Goal Ach. | | | | ✓ | | | | ✗ | | | | ✗ |

Table 9: Results of Two-Tailed Wilcoxon Signed Rank Tests comparing different sets of variables as integers in the MIP (using a layer-weighting of $k = 1.1$). ✓ indicates significance ($p = 0.05$), colour indicates the better performer (faster, fewer nodes expanded or shorter plans).

| | Time Taken | | | | Nodes Expanded | | | | Plan Length | | | |
|---|---|---|---|---|---|---|---|---|---|---|---|---|
| | First Layer Acts | Prop. Goal Ach. | Num. Goal Ach. | All Ints | First Layer Acts | Prop. Goal Ach. | Num. Goal Ach. | All Ints | First Layer Acts | Prop. Goal Ach. | Num. Goal Ach. | All Ints |
| Minimal Ints | ✓ | ✓ | ✓ | ✓ | ✗ | ✗ | ✗ | ✗ | ✗ | ✗ | ✗ | ✗ |
| First Layer Acts | | ✓ | ✓ | ✓ | | ✗ | ✓ | ✓ | | ✗ | ✗ | ✗ |
| Prop. Goal Ach. | | | ✓ | ✓ | | | ✗ | ✓ | | | ✗ | ✗ |
| Num. Goal Ach. | | | | ✓ | | | | ✗ | | | | ✗ |

Table 10: Results of Two-Tailed Wilcoxon Signed Rank Tests comparing different sets of variables as integers in the MIP (using a layer-weighting of $k = 3$). ✓ indicates significance ($p = 0.05$), colour indicates the better performer (faster, fewer nodes expanded or shorter plans).

To confirm significance of these observations, we applied Two-Tailed Wilcoxon Signed Rank Tests, the results of which are in Tables 9 and 10, for $k = 1.1$ and $k = 3$, respectively. In both cases, there is a consistent increase in the time taken to solve problems as the proportion of integral action variables is increased: this difference is significant in every case other than when using $k = 1.1$ and comparing 'First Layer Actions' to 'Propositional Goal Achievers'. There is no significant difference in plan length between any pair of configurations. The results for nodes are somewhat less clear. It does not appear that using 'Minimal Integers' leads to expansion of a significantly different number of nodes,

| Domain | Minimal Ints | | First Layer Actions | | All Ints | |
|---|---|---|---|---|---|---|
| | Build (ms) | Solve (ms) | Build (ms) | Solve (ms) | Build (ms) | Solve (ms) |
| Market Trader(20) | 26.4 | 27.7 | 27.8 | 52.1 | 27.2 | 55.3 |
| Hydro Power (21) | 8.8 | 1.8 | 9.2 | 6.5 | 10.8 | 6.2 |
| Pathways Metric (30) | 58.5 | 271.2 | 64.9 | 834.0 | 65.4 | 3468.9 |
| Sugar (16) | 24.9 | 33.8 | 26.1 | 46.6 | 24.7 | 329.2 |
| Settlers (9) | 282.3 | 96.1 | 453.8 | 201.6 | 295.2 | 204.7 |
| Settlers Num. Carts (16) | 399.4 | 101.6 | 520.9 | 462.8 | 408.8 | 222.1 |
| MPrime (28) | 78.8 | 4.8 | 75.6 | 6.3 | 78.9 | 6.1 |
| Rovers Numeric (11) | 57.7 | 1.7 | 42.4 | 2.1 | 59.0 | 2.8 |
| Average | 117.1 | 67.3 | 152.6 | 201.5 | 121.3 | 536.9 |

Table 11: Time spent in building and solving LP varying which variables are integer in the MIP; numbers shown with domains indicate how many problem instances were solved and used in computing the reported average

when compared to any other configuration. In part this is due to a limitation of the tests — only pairwise-solved problems can be included, so the difference in coverage is not reflected in the data analysed in the test. Considering the other configurations, though, 'All Integers' evaluates fewer nodes than 'First Layer Actions' and 'Propositional Goal Achievers', regardless of which weight is used, but cannot be shown to expand fewer nodes than 'Numeric Goal Achievers', the configuration that offered better coverage in Table 8.

Although we have seen that increasing the number of integer variables makes the planner slower, it is somewhat surprising that the amount by which the planner is made slower is not greater than it is. In theory we would expect that making all variables integer would dramatically decrease the performance of the planner, however, it appears this is not so. To investigate this further we consider how long was spent building and solving the LP in each state for each configuration. Table 11 gives an indication of why we are seeing this surprising result: in 5 out of 8 domains the solving time of the LP is less than the building time (for the All Ints configuration), showing again that a key bottleneck in using the LP solver is, in fact, in building the LP. Building time is, of course, not likely to vary between configurations (the only variation is due to different states being expanded) and compared to this the solving time is small. In two of the three remaining domains, Pathways and Sugar, we see some increase in the cost of solving the LP when first-layer actions are made integral, and a much larger increase when all actions are made integral, this is the pattern that we expected. In the final domain, Market Trader, the main increase occurs when first-layer action variables are integral: the structure of this domain means that integral first layer actions often causes the variables in other layers to also become integral. Solving times does, however, remain within an order of magnitude of building time, so the overhead is not particularly large in this domain compared to the previous two.

## 8.6 Including the Numeric Goal Conjunct

In Section 7.1 we discussed the possibility of including the entire numeric goal conjunct for the problem in the LP. As well as theoretically increasing the ability to detect dead-ends — by insisting all goals are attainable at the same time, rather than individually — this also allows arbitrary LNF goals to be used, as found in domains such as Pathways. In this section, we will investigate the impact of this extension. In particular, we explore whether the use of a numeric-goal-checking LP including the numeric goal conjunct improves or worsens performance. To ensure that there is no goal-checking LP in the case where the numeric goal conjunction is not being used, for these tests, in both configurations, we disable the inclusion of propositional goals and landmarks in the LP. To gain insights into how the impact of a numeric-goal-checking LP is affected by the choice of action-variable layer-weighting schemes, we consider two: $k = 1.1$ and $k = 3$.

| Domain | With Num.Goal $(k = 3)$ | Without Num. Goal $(k = 3)$ | With Num. Goal $(k = 1.1)$ | Without Num. Goal $(k = 1.1)$ |
|---|---|---|---|---|
| Market Trader | 20 | 20 | 20 | 20 |
| Hydro Power | 27 | 27 | 29 | 30 |
| Pathways Metric | 30 | 0 | 28 | 0 |
| Sugar | 18 | 14 | 9 | 14 |
| Settlers | 13 | 14 | 12 | 10 |
| Settlers Num. Carts | 19 | 18 | 21 | 22 |
| MPrime | 28 | 29 | 28 | 29 |
| Rovers Numeric | 13 | 13 | 13 | 13 |
| Total (Excl. Pathways) | 138 | 135 | 132 | 138 |
| Total | 168 | 135 | 160 | 138 |

Table 12: Coverage varying whether the Numeric Goal conjunct is included in the LP (with) or not (without)

The coverage results for running LP-RPG with and without the numeric goal conjunct are shown in Table 12. Looking at the results excluding the Pathways domain (which can only be solved if the numeric goal conjunct is included, since the goals are expressed as an arbitrary LNF), one can make two immediate observations: the use of the numeric goal conjunct *improves* performance with actions weighted as $k = 3$, solving 3 additional problems, but its use *worsens* performance with actions weighted as $k = 1.1$, solving 6 fewer problems. This difference in impact is an interesting consequence of the relationship between the RPG structure and the LP:

- In the no-goal-conjunct case, if a goal appears at fact layer $l$, the LP used to meet this goal is $LP(l)$ — the LP containing the actions up to action layer $l$ (Algorithm 4, line 22). This favours the earlier actions in the RPG, precluding any actions after layer $l$ from being used.

- In the numeric goal-conjunct case, the LP is extended until the layer $l'$ at which, first, all goals appear, and second, an LP constrained to meet all the goals, using the actions up to layer $l'$, can be satisfied. For individual goals, this point may be later than the layer $l$ at which they first appeared, in which case actions that are in $l'$ but not $l$ can be chosen to meet these goals, whereas in the previous case they could not.

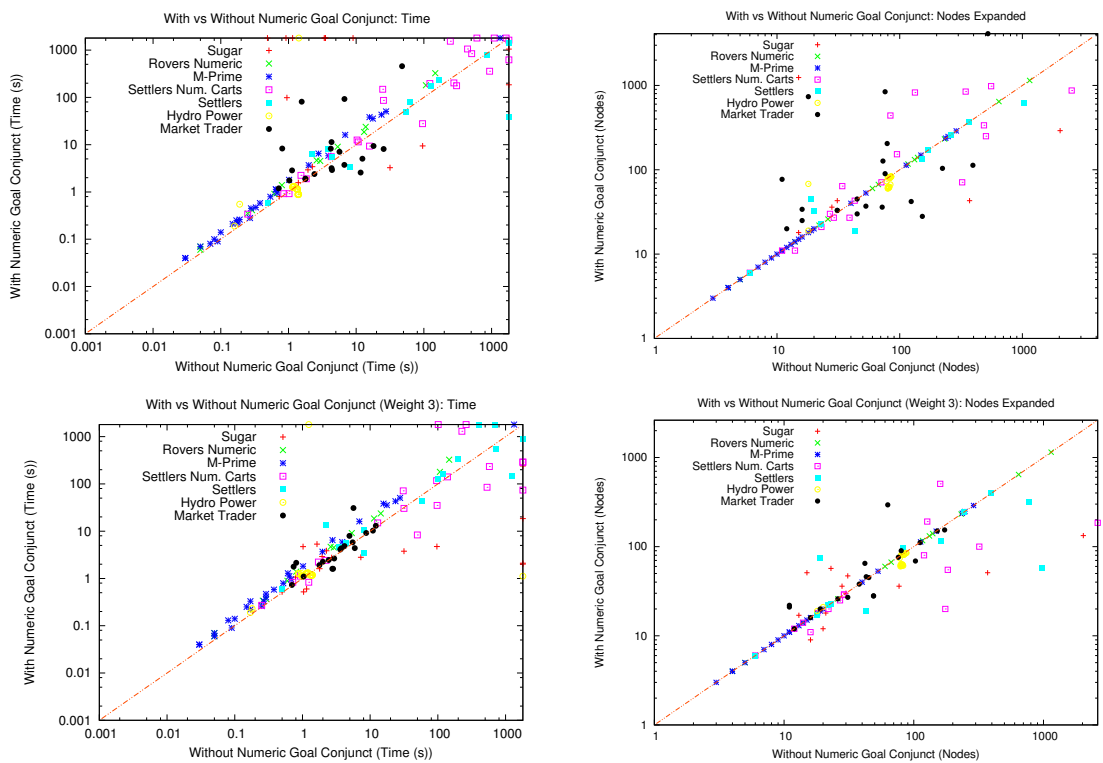

Figure 11: Including the numeric goals in the LP

It is this difference in the actions that are available to support a given goal that makes the action-variable weighting scheme important. The use of $k = 3$ rather than $k = 1.1$ means that while actions added between layer $l$ and layer $l'$ (inclusive) are available to meet a goal that first appeared at layer $l$, the objective function leads to a preference to use earlier actions.

The coverage results in these experiments do not give an unequivocal picture of the impact of this feature on performance. Figure 11 shows scatterplots of the time taken to solve problems, and the number of nodes evaluated, when the numeric goal-conjunct is used or not, for each of $k = 1.1$ and $k = 3$. There appears to be a general trend for the inclusion of the numeric goal conjunct to reduce the number of nodes evaluated. In the case of weighting by $k = 3$, a Two-Tailed Wilcoxon Signed Rank Test confirms that the use of a numeric goal conjunct reduces the number of nodes evaluated ($p = 0.05$). Using $k = 1.1$ suggests a similar trend, but the results are not significant.

With both weighting schemes, the use of the numeric goal conjunct introduces a small but statistically significant time overhead. This is due to the additional time taken to evaluate each state: the RPG must be extended to the point at which all goals can be satisfied together, rather than individually. Whether this pays off, i.e. whether the reduction in nodes evaluated is sufficient to offset this, depends on the domain and appears to be correlated with the extent to which the numeric goals interact. At one extreme, in the Rovers domain, all the goals are propositional and there is no difference in performance. On the other hand, in the Sugar domain, and the Settlers domain when numeric carts are used, it is beneficial. Both of these domains concern production and reprocessing of raw materials, in one form or another, leading to interaction between goals. For instance, a unit of a resource may be sufficient to satisfy goals individually, but additional production may be required to support them both. In these cases use of the numeric goal conjunct improves time performance. Inclusion of the numeric goal conjunct has no significant impact on the length of plans produced (according to a Two-Tailed Wilcoxon Signed Rank Test).

To summarise the results in this section, the main benefit from the use of the numeric goal conjunct is to be able to extend the expressivity of the planner to domains where the goals are written using arbitrary LNF. The success of the approach in other domains varies. In terms of coverage, whether or not it is better to use the numeric goal conjunct on the evaluation domains depends on the weighting scheme. When using the layer-weighting scheme, $k = 3$, the inclusion of the numeric goal conjunct is slightly beneficial and LP-RPG is therefore set to use this configuration by default.

## 8.7 Including Propositions in the LP

In the previous section, we observed that the inclusion of the numeric goal conjunct in a goal-checking LP has variable impact on performance, depending on the weighting scheme used. Perhaps a more interesting use of a goal-checking LP is when using the LP to meet propositional goals, and landmarks, as described in Section 6.1. To evaluate this technique we consider four configurations:

1. No propositions: using a goal-checking LP containing only the numeric-goal conjunct.

2. Propositional goals: as in previous case, but also including the propositional goals in the goal-checking LP.

3. Landmarks: as in previous case, but also including landmarks.

4. All propositions: as in previous case, but also constrained to ensure that if an action variable is non-zero, actions are added to meet its propositional preconditions (as described in Section 6.3).

These form a spectrum, from the case in which no information about propositions is included in the LP at all, to the last in which the goal-checking LP must not only meet the propositional goals, but also the preconditions of the actions chosen to do so. We consider two layer-weighting schemes ($k = 1.1$ and $k = 3$), and action variables for actions in the first action layer are integral.

The coverage results for $k = 1.1$ are shown in Table 13 and those for $k = 3$ in Table 14. As can be seen, in both cases, a general pattern emerges: coverage improves up to and including the configuration in which landmarks are included in the LP, but then declines in the final 'All Propositions' configuration. Including all propositions appears, however, to remain better than including no propositions at all.

Scatterplots illustrating the time taken and the number of nodes evaluated when solving problems are shown in Figures 12 and 13 (for weights $k = 1.1$ and $k = 3$, respectively).

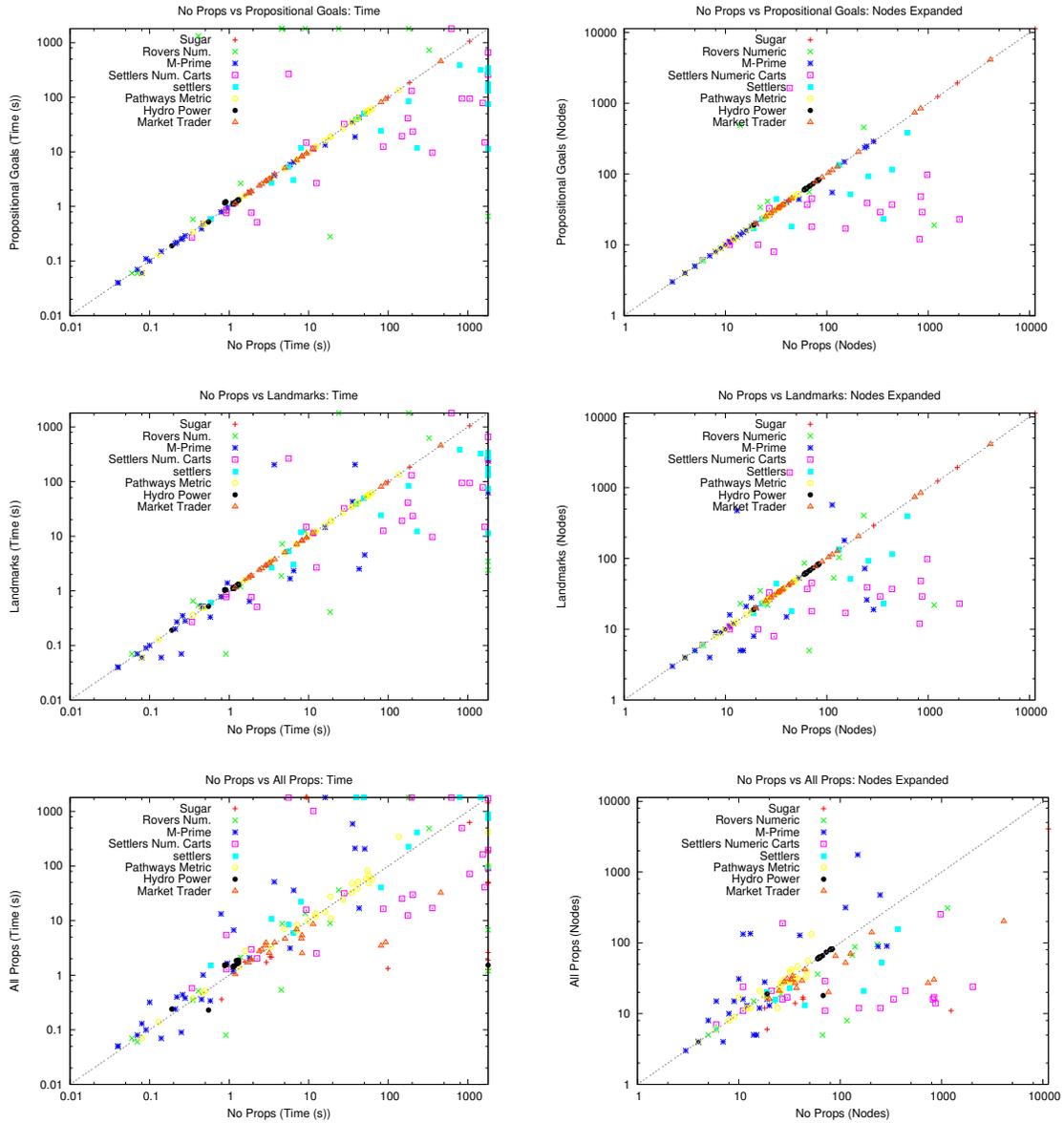

Figure 12: Varying which propositions are included in the LP ($k = 1.1$)

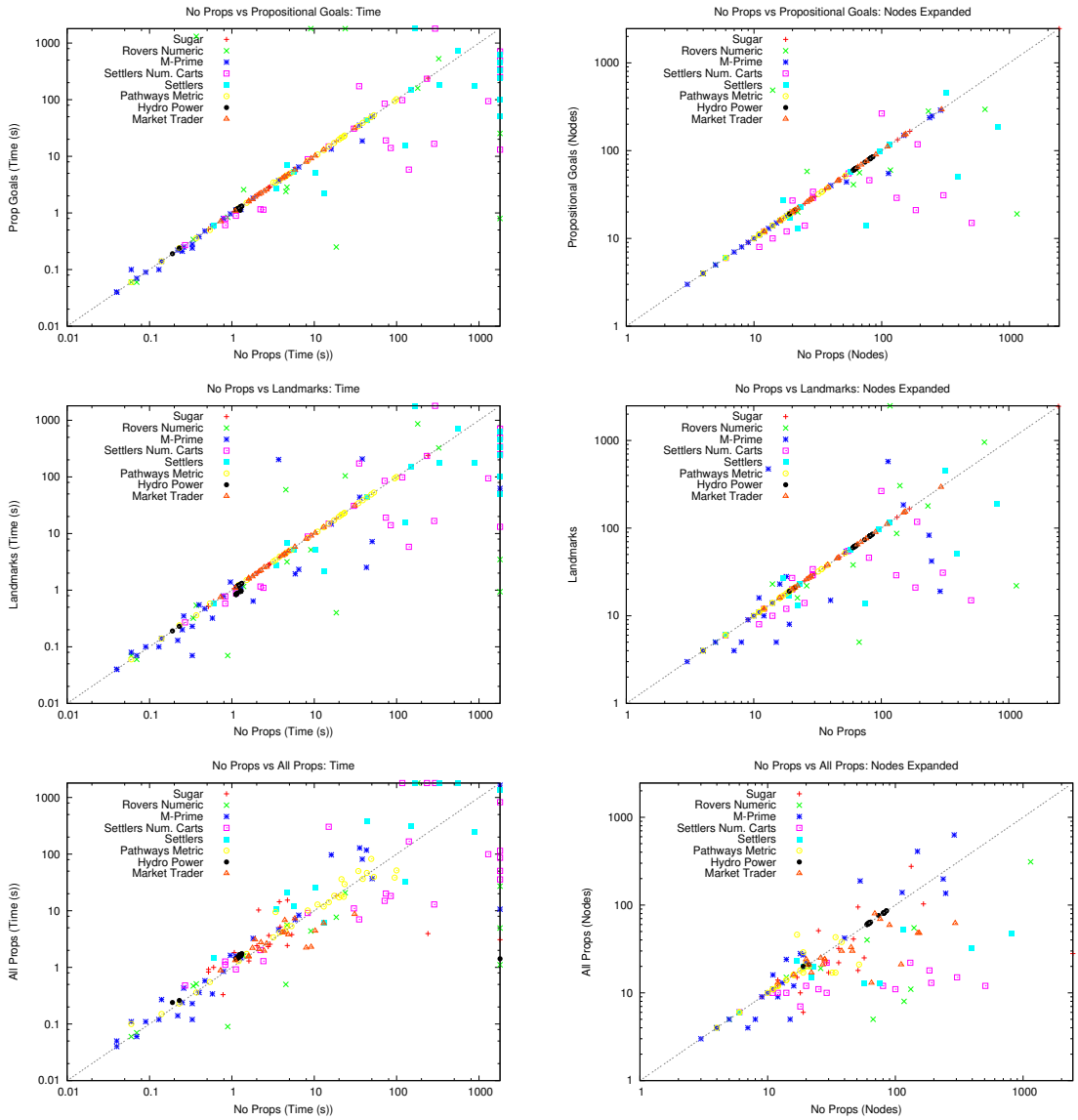

Figure 13: Varying which propositions are included in the LP ($k = 3$)

| Domain | No Props | Prop Goals | Landmarks | All Props |
|---|---|---|---|---|
| Market Trader | 20 | 20 | 20 | 19 |
| Hydro Power | 29 | 29 | 29 | 30 |
| Pathways Metric | 28 | 28 | 28 | 30 |
| Sugar | 9 | 9 | 9 | 14 |
| Settlers | 12 | 19 | 19 | 11 |
| Settlers Numeric | 21 | 22 | 22 | 21 |
| MPrime | 28 | 28 | 30 | 27 |
| Rovers Numeric | 13 | 9 | 13 | 15 |
| Total | 160 | 164 | 170 | 167 |

Table 13: Coverage when varying which propositions are included in the LP (using $k = 1.1$)

| Domain | No Props | Prop Goals | Landmarks | All Props |
|---|---|---|---|---|
| Market Trader | 20 | 20 | 20 | 20 |
| Hydro Power | 27 | 27 | 27 | 28 |
| Pathways Metric | 30 | 30 | 30 | 30 |
| Sugar | 18 | 18 | 18 | 19 |
| Settlers | 13 | 18 | 18 | 11 |
| Settlers Numeric | 19 | 23 | 23 | 21 |
| MPrime | 28 | 28 | 30 | 30 |
| Rovers Numeric | 13 | 13 | 15 | 14 |
| Total | 168 | 177 | 181 | 173 |

Table 14: Coverage when varying which propositions are included in the LP (using $k = 3$)

Each scatterplot compares one of the configurations 2–4 in the enumerated list above to configuration 1 (no propositions). Comparing the 'Propositional Goals' and 'Landmarks' configurations suggests that the inclusion of Landmarks leads to more problems being solved more quickly compared to the 'No propositions' configuration. This is particularly evident in the Settlers, Settlers Numeric Carts, MPrime and Rovers Numeric domains. Whether using Landmarks offer gains over Propositional Goals alone depends on the structure of the problems. In the two Settlers variants, including propositional goals (such as (`connected-by-rail p1 p2`) or (`has-ironworks p3`)) necessitates the use of specific actions ((`build-rail p1 p2`) or (`build-ironworks p3`), respectively). Since these actions consume numeric resources, including propositional goals is sufficient to cause the LP to introduce the production actions necessary to support the resource consumption. In Rovers, the goals are to have communicated data generated from rock samples, soil analyses and images. The actions adding these facts consume a number of units of energy, so there is some benefit from including propositional goals in the LP. However, these actions also have propositional preconditions: for instance, for a rover to communicate rock data, it must have taken a sample of the rock. These propositional preconditions will be included as landmarks in the LP, thereby forcing the production of not only the energy needed to communicate the requisite data, but also the energy to have acquired it. This leads to the inclusion of additional recharge actions or, if no suitable actions are available, better dead-end detection. A similar situation arises in MPrime: the actions that achieve the goal facts do not consume any resources, but the actions that achieve the (landmark) preconditions of these actions do, so including the landmarks improves informedness of the heuristic.

Two-Tailed Wilcoxon Signed Rank Tests — shown in Tables 15 and 16 — confirm ($p = 0.05$) that with either $k = 1.1$ or $k = 3$ the use of Propositional Goals improves over No Propositions, and the use of Landmarks improves over Propositional Goals. Indeed, with $p = 0.05$ we have a total-ordering on time-performance for $k = 3$:

$$\text{No Propositions} \prec \text{Propositional Goals} \prec \text{Landmarks}$$

The results show similar behaviour for numbers of nodes evaluated (the right-hand columns of Figures 12 and 13) although not all of the differences are statistically significant. Wilcoxon tests also show that with neither $k = 1.1$ or $k = 3$ is the difference between Propositional Goals and Landmarks statistically significant. Note, however, that using Landmarks allows more problems to be solved. The comparisons for No Propositions, Propositional Goals and Landmarks, with weight $k = 1.1$, are significant, the latter two both improving upon the former. With weight $k = 3$ the same tests are inconclusive, although using Landmarks allows 13 additional problems to be solved (recall that the comparisons are restricted to the problems solved by both variants in the comparison).

| | Time Taken | | | Nodes Expanded | | | Plan Length | | |
|---|---|---|---|---|---|---|---|---|---|
| | Prop Goals | Land-marks | All Props | Prop Goals | Land-marks | All Props | Prop Goals | Land-marks | All Props |
| No Props | ✓ | ✓ | ✗ | ✓ | ✓ | ✓ | ✗ | ✗ | ✓ |
| Prop Goals | | ✓ | ✓ | | ✗ | ✓ | | ✗ | ✓ |
| Landmarks | | | ✓ | | | ✓ | | | ✓ |

Table 15: Results of Two-Tailed Wilcoxon Signed Rank Tests comparing inclusion of different propositions in the LP (using layer-weighting $k = 1.1$). ✓ indicates significance ($p = 0.05$), colour indicates the better performer (faster, fewer nodes expanded or shorter plans).

| | Time Taken | | | Nodes Expanded | | | Plan Length | | |
|---|---|---|---|---|---|---|---|---|---|
| | Prop Goals | Land-marks | All Props | Prop Goals | Land-marks | All Props | Prop Goals | Land-marks | All Props |
| No Props | ✓ | ✓ | ✗ | ✓ | ✗ | ✓ | ✗ | ✗ | ✓ |
| Prop Goals | | ✓ | ✓ | | ✗ | ✓ | | ✗ | ✓ |
| Landmarks | | | ✓ | | | ✓ | | | ✓ |

Table 16: Results of Two-Tailed Wilcoxon Signed Rank Tests comparing inclusion of different propositions in the LP (using layer-weighting $k = 3$). ✓ indicates significance ($p = 0.05$), colour indicates the better performer (faster, fewer nodes expanded or shorter plans).

| Domain | No Props | | Landmarks | | All Props | |
|---|---|---|---|---|---|---|
| | Build (ms) | Solve (ms) | Build (ms) | Solve (ms) | Build (ms) | Solve (ms) |
| Market Trader (20) | 27.8 | 52.1 | 28.0 | 52.3 | 29.7 | 74.5 |
| Hydro Power (27) | 9.3 | 6.6 | 9.8 | 6.6 | 11.7 | 5.5 |
| Pathways Metric (30) | 64.9 | 834.0 | 67.0 | 833.3 | 68.6 | 905.5 |
| Sugar (18) | 27.2 | 47.5 | 27.3 | 47.5 | 39.9 | 82.3 |
| Settlers (8) | 436.4 | 145.5 | 421.6 | 137.9 | 563.9 | 3153.0 |
| Settlers Num. Carts (15) | 204.5 | 434.8 | 252.5 | 133.0 | 244.2 | 2253.8 |
| MPrime (28) | 75.6 | 6.3 | 78.9 | 5.7 | 91.2 | 24.5 |
| Rovers Numeric (11) | 42.4 | 2.1 | 56.3 | 2.0 | 76.3 | 31.2 |
| Average | 111.0 | 191.1 | 117.7 | 152.3 | 140.7 | 816.3 |

Table 17: Time spent building and solving the LP, varying which propositions are included; numbers shown with domains indicate how many problem instances were solved and used in computing the reported average

The results shown at the bottom of Figures 12 and 13 indicate that the 'All Propositions' configuration has less consistent performance.

We can show, however, ($p = 0.05$) that with $k = 1.1$ or $k = 3$, the All Propositions configuration takes longer to solve problems than both the Propositional Goals configuration and the Landmarks configuration. It is perhaps surprising that it is feasible to consider including propositions in the LP in this manner, given the results reported by van den Briel et al. (2007). The key difference is that we are disregarding delete effects, so if a fact is needed as a precondition it need only be added at most once. In the work reported by van den Briel et al., however, in cases where a fact is required as a precondition but also deleted, the delete effect must be balanced by an equivalent number of add effects (less one if the fact was true initially).

Looking at the number of nodes evaluated when using the All Propositions configuration we find an important result: as shown in the bottom right of Figures 12 and 13, whichever weight is used, All Propositions tends to expand fewer nodes. Furthermore, All Propositions expands fewer nodes than all three of the other configurations and is the overall best configuration in terms of nodes expanded (significant result, $p = 0.05$). Unfortunately, the overhead associated with each node is higher and this results in poorer coverage and time performance. This result indicates that numeric–propositional separation used in the LP-RPG heuristic is a sensible trade-off, with the use of the RPG to meet propositional preconditions sacrificing some performance in terms of nodes expanded in exchange for a reduction in time taken to solve problems.

Table 17 shows the increase in costs associated with solving LPs including propositions. These results are taken only from problems solved by all three configurations, although it remains the case that the configurations will not expand exactly the same states in solving the same problems. In three domains — Hydro Power, MPrime and Rovers Numeric — the cost of building LPs dominates the cost of solving them, so there are no significant decreases in performance in these domains. Of course, the cost of building LPs increases as more propositions are included since more variables are required. In the Settlers domain there is an order of magnitude increase in the time spent solving LPs, indicating that including all propositions in the LPs makes them more difficult to solve. Sometimes the more informed search guidance gleaned from this information allows the planner to find solutions expanding far fewer nodes.

Adding landmarks to the LP appears to make solving the LP slightly faster: this is partly a side effect of the landmarks configuration needing to solve fewer LPs per state, since it uses one LP to meet all the goals instead of one LP per goal. The All Propositions configuration often only needs to solve one LP in the solution extraction phase. However, the size and difficulty of this LP means that any benefits of solving fewer LPs (in terms of time taken) are negated.

Tables 15 and 16 show the results of statistical tests comparing the lengths of the plans found by the various configurations. The only significant results are that, regardless of the weight used, All Propositions finds shorter plans than the other configurations. If all propositions are included in the LP, then typically only a single LP call is made during solution extraction, simultaneously achieving all goals and action preconditions. In the other configurations, first, an LP call is made for the goals, and then, for each action $A$ added to support a propositional precondition of an action implied by the solution to this LP, the numeric preconditions of $A$ are met by another LP call. Thus, there can be several LP calls rather than just one. By fragmenting the production of a relaxed plan in this way, the efficacy of the relaxed plan is eroded. As an example, plan lengths are improved in the two variants of Settlers. Here, the production of some resources requires building infrastructure: a sawmill is required to refine timber into wood, and so on. If two units of wood are to be made, then if the LP has no knowledge of propositional preconditions, there is no difference in the LP between building two units of wood at location $A$, or one unit at location $B$ and one at location $C$ — the need to build one or two sawmills, depending on which option is chosen, will only be discovered when actions are then chosen to meet the propositional preconditions of the actions required by the solution to the LP. If the LP includes All Propositions, then in cases such as this, there *is* a difference between building one sawmill and two, so the LP will prefer the single-sawmill solution, ultimately leading to better search guidance. Thus, All Propositions produces plans that are significantly shorter, in domains where the fact that propositions are outside the LP disguises the true costs of the action choices in the solution to an LP.

## 8.8 Propositional Resource Analysis

At the end of Section 6.4, we identified conditions under which it is possible to turn propositions that model resources into an equivalent numeric representation. For our purposes, with LP-RPG, this would allow the resource, within the heuristic, to be managed by the LP rather than the RPG. To this end, we use three domains containing propositional stacks that represent resources, and evaluate how this encoding affects the performance of LP-RPG, and other numeric planners. The five planners evaluated are MetricFF, LPG–td, LP-RPG (using the analysis that translates propositional resource stacks into an equivalent numeric representation as described in Long & Fox, 2000), LP-RPG with this analysis disabled, and LP-RPG-FF. The three domains we use are:

| Domain | MetricFF | LPG–td | LP-RPG-FF | LP-RPG | |
|---|---|---|---|---|---|
| | | | | No Anal. | Anal. |
| Settlers Prop. Timber | 7 | 4 | 8 | 5 | 16 |
| Settlers Prop. Carts | 4 | 4 | 10 | 13 | 22 |
| MPrime Prop. | 29 | 30 | 29 | 29 | 30 |
| Total | 40 | 38 | 68 | 47 | 47 |

Table 18: Coverage with and without propositional resource analysis

- A variant of the IPC3 Settlers domain, where the amount of timber at a given location (or in a given vehicle) is represented by a stack of propositions, in the range `n0` to `n10`.

- A variant of the Settlers 'Numeric Carts' encoding, where the number of carts at a given location is represented by a stack of propositions.

- The MPrime domain from IPC1, which in its encoding represents fuel level as a stack of propositions.

In the first two of these, the propositional encoding enforces a limit of 10 on the amount of timber (respectively, number of carts) at a given location. This is a necessary limitation forced by action grounding. In a numeric representation, decreasing one resource and increasing another can be done by a single ground action, with appropriate numeric effects and conditions. The ground action does not need to stipulate the precise levels of the resource before and after the operation, so long as the limits on the resource levels are respected. In the propositional case, however, one ground action is needed for each pair of discrete levels of the two resources, with parameters to the action stipulating the pre- and post-values of each resource. The level of 10 was chosen to avoid placing overly restrictive bounds on the resource levels, while creating a manageable number of ground actions. This issue of grounding also accounts for the limited choice of domains: a propositional stack can only be used to represent resources that can take a modest range of discrete values. In the Rovers domain, for example, the set of reachable energy levels for each rover is in the range $[0, 80]$. In the Market Trader domain, the amount of money that could be held at a given time is a real value with one decimal place (this is a consequence of the choice of problem files), greater than or equal to zero.

Results for these domains are shown in Table 18. Comparing, first, LP-RPG without analysis to LP-RPG with analysis, we can see that encoding the resources in the LP as numbers grants a consistent improvement in performance. Figure 14 shows that the time taken to find solutions is similarly improved, perhaps most strikingly in the second Settlers variant (with propositional cart levels).

While MetricFF and LPG–td perform well in the MPrime domain, in the two Settlers encodings, LP-RPG with the resource analysis performs markedly better, with the other planners solving only a handful of problems. This contrasts with their earlier results, shown in Table 2, where LPG–td in particular performed well on the Settlers domain. It is an interesting contrast to see that here, the Propositional Timber variant — derived from the IPC 3 model on which LPG–td performs well — leads to considerably worse performance.

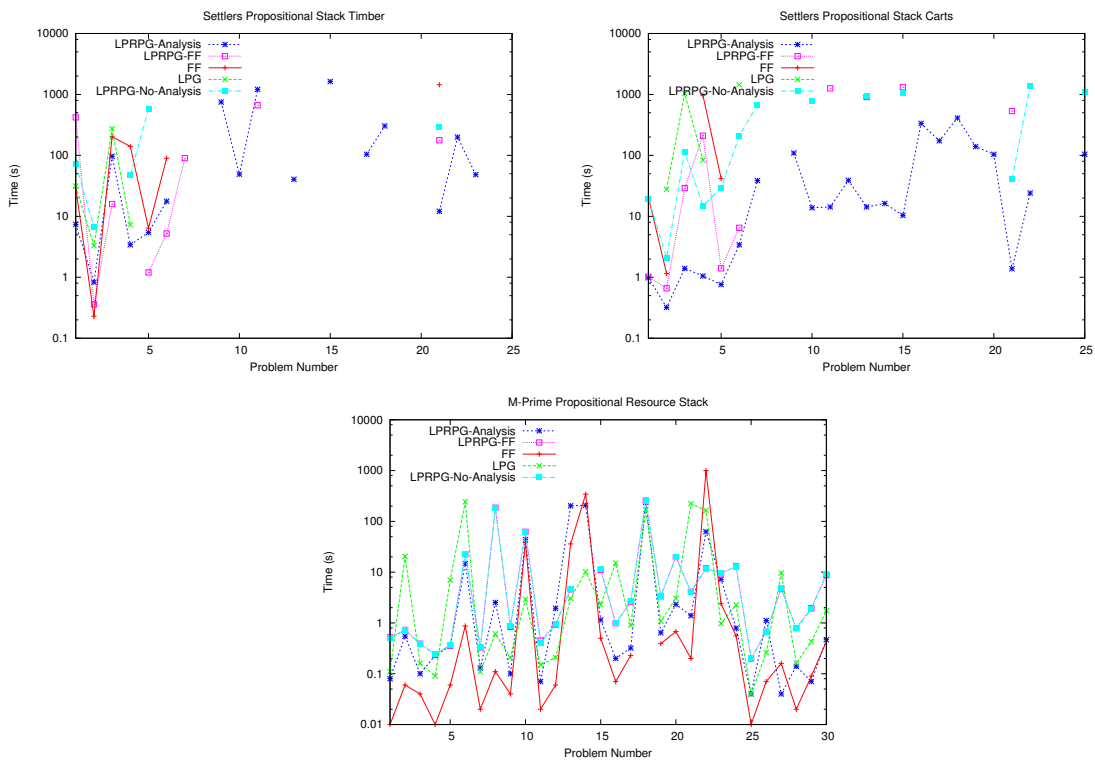

Figure 14: Using propositional resource analysis

This supports the role of resource analysis in allowing LP-RPG to be robust to which of a number of almost-equivalent domain encodings is used, with mixed propositional and numeric encodings of resource levels. We observed no significant change in the length of plans produced by LP-RPG with or without propositional resource analysis.

## 9. Conclusions and Future Work

Most modern planning systems are ineffective at reasoning with numbers. However, managing complex numeric interactions is an important part of driving AI planning towards future real-world application. In this paper we have shown that, by using a linear program to model numeric resource flows, the ability of planners to reason with domains involving such complex numeric interactions can be greatly improved.

The key contribution is the separation of the heuristic search control into a relaxed planning graph, based on delete-relaxation, and a linear program that allows exact reasoning about numeric constraints and relaxes action ordering.

We have explored how different configurations of the heuristic, in which we put more or less information in to the LP, impact on the performance of the planner as a whole. An exploration of different LP solvers reveals that they are more or less efficient at handling various combinations of constraints. We found that, while LPSolve and CLP, in conjunction with the limited version of LP-RPG published in 2008 (Coles et al., 2008), can solve simple problems quickly, CPLEX coupled with the full power of the extended LPRPG is needed to handle the most complex test instances.

Our work so far has focussed on developing search control methods that can perform well on numeric planning problems with a particular character: the producer-consumer behaviour we define in Sectionsect:prodcondefinition. Although we believe that this is a common behaviour, in practice, numeric domains exhibit a range of other behaviours. There are several possible ways to exploit the LP-RPG approach in these domains. One is to use the approach on those actions that conform to the constraints of producer-consumer behaviour, while pushing other numeric behaviour into a metric RPG in the same way that we currently handle propositional goals and preconditions in a separate RPG. This would yield the benefits of potentially better estimates for reachable ranges and action use costs for those parts of the domain that we can express as producer-consumer behaviour. More challenging is to consider how other behaviours can be relaxed into producer-consumer behaviour to obtain useful heuristic information. For example, actions with production effects that vary could be encoded as a family of producers of increasing capability, discretising the range of production options and introducing them into the reachability analysis as their increased production capabilities become available. In general, the relaxations must make reachability at least as permissive as actual reachability (that is, an action must be applicable in the relaxed reachability analysis at least as early as the action is actually reachable) and the relaxed plan extraction should minimise the estimated cost to goal effectively (this is more difficult because the relaxed plan is not optimal). Within these constraints, we believe that the use of LP approximations can provide a tool for tackling a wider range of behaviours than those we explore in this paper.

A further exciting challenge for our future work is to integrate LP-RPG with a method for optimising plans according to a given objective function. The recent 2008 and 2011 planning competitions highlighted the importance of optimising planning with their emphasis on solution quality. This development is, however, non-trivial: the challenges lie in the integration of cost optimisation between the LP and the RPG as well as in deciding how to use a heuristic that trades off goal distance for quality during search. A first step in this direction was accomplished by Radzi in her PhD thesis (Radzi, 2011), but the advances in numeric planning described in this paper open up many possibilities for extending that initial work.